\documentclass[letterpaper]{article}
\usepackage[preprint]{aaai2027}
\usepackage[hyphens]{url}
\usepackage{graphicx}
\usepackage{natbib}
\usepackage{caption}
\usepackage{float}
\usepackage{booktabs}
\usepackage{array}
\usepackage{amsmath}
\usepackage{amssymb}
\usepackage{xspace}
\usepackage[most]{tcolorbox}

\newcommand{\method}{JustLLMGRPO\xspace}
\makeatletter
\newcommand{\correspondingdagger}{%
  \ifx\footnote\relax
    \corresponding
  \else
    \setcounter{footnote}{1}%
    \begingroup
      \renewcommand{\@makefnmark}{}%
      \corresponding
    \endgroup
  \fi
  \raisebox{0.75ex}{\fontsize{7}{7}\selectfont\ensuremath{\dagger}}}
\makeatother
\newcolumntype{L}[1]{>{\raggedright\arraybackslash}p{#1}}
\definecolor{PromptPurple}{HTML}{6C4CCF}
\definecolor{PromptBlue}{HTML}{4776C5}
\definecolor{PromptCyan}{HTML}{2A9D9F}
\definecolor{PromptInk}{HTML}{26344D}
\definecolor{PromptBackground}{HTML}{F7F8FC}

\title{JustLLMGRPO: Radiographic Control for Chest X-Ray Generation}
\author{
Pengxiang Cai,
Xiaohan Li,
Anglin Liu,
Qingyuan Zeng,
Zexun Li,
Jintai Chen\correspondingdagger
}
\affiliations{
The Hong Kong University of Science and Technology (Guangzhou)\\
jintaiCHEN@hkust-gz.edu.cn}

\begin{document}

\maketitle

\begin{abstract}
Text-conditioned chest X-ray generation aims to synthesize realistic radiographs that faithfully depict specified findings. Existing work has primarily improved quality by updating image generators, implicitly treating prompts as fixed after CXR-domain adaptation. We show that this generator-centric view leaves a substantial optimization dimension underexplored. With a CXR-adapted Sana generator frozen, one-pass reformulation by an unmodified LLM reduces RadDINO-FID from 54.225 to 27.572. Prompt analysis shows that the LLM suppresses temporal comparisons, uncertainty, and other non-renderable report content while emphasizing visible radiographic findings. However, unconstrained reformulation reduces BioViL-T alignment with source prompts from 0.695 to 0.609. We therefore introduce \method, which applies standard Group Relative Policy Optimization (GRPO) only to the LLM prompt policy while keeping Sana frozen. Group-relative radiology-aware image feedback retains visual focus while preserving source-prompt alignment. On CheXGenBench, \method reduces RadDINO-FID to 26.780, a 50.6\% improvement over direct prompting, while maintaining alignment (0.696 versus 0.695). It also achieves state-of-the-art distribution coverage and downstream classification utility. These results show that substantial performance can remain latent in how radiographic information is expressed to an adapted generator. Code is publicly available at \url{https://github.com/pxcai/JustLLMGRPO}.
\end{abstract}

\section{Introduction}

Text-conditioned chest X-ray (CXR) generation aims to synthesize realistic radiographs that faithfully depict specified findings. Existing work has predominantly improved the generator through domain adaptation, specialized conditioning, and clinically informed reinforcement learning~\citep{weber2023cascaded,bluethgen2025visiona,lee2024visionlanguage,lee2024llmcxr,han2024advancing,moris2024adapted,shentu2024cxrirgena,black2024training,fan2023dpok,prabhudesai2024aligning}. These approaches have strengthened radiograph realism and clinical control, establishing generator-side adaptation as the dominant route to further performance gains.

Across this trajectory, prompts are generally treated as fixed once a generator has been adapted to the CXR domain. Subsequent improvements are sought by changing its architecture, parameters, or learned conditioning representations. This practice implicitly assumes that the remaining bottleneck lies within the generator. Yet the conditioning text determines which findings reach the renderer and how they are expressed. Whether an adapted generator can improve without further parameter updates therefore remains underexplored.

This question is especially important because CXR prompts are commonly derived from radiology reports. Reports are clinical communication documents rather than literal image descriptions. They mix current findings with prior comparisons, uncertainty, examination context, and longitudinal discourse. A single CXR, however, depicts one static visual state, so much of this information cannot be rendered directly. When used as prompts, non-renderable content can dilute the signal from visible findings. This mismatch suggests that an LLM could translate report-derived prompts into image-realizable conditions without changing the generator.

\begin{figure}[t]
    \centering
    \includegraphics[width=\columnwidth]{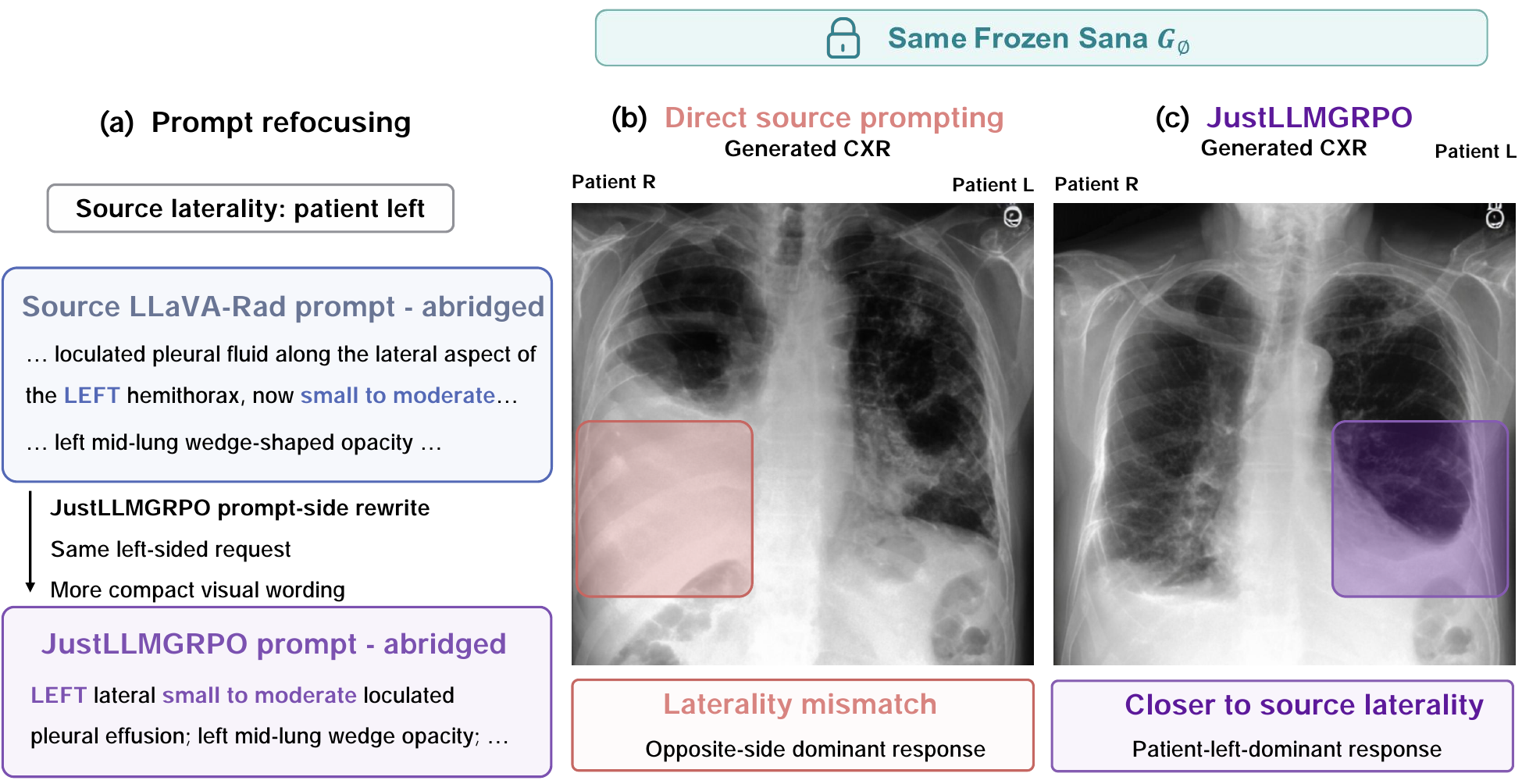}
    \caption{\textbf{Prompt optimization refocuses a frozen CXR generator.} With Sana frozen, the source prompt produces an incorrect CXR with right-dominant effusion, whereas the \method prompt restores the requested laterality.}
    \label{fig:discovery}
\end{figure}

Prior prompt-optimization studies show that language models can adapt prompts for frozen text-to-image generators through rewriting, expansion, or iterative visual correction. These methods mainly enrich short generic descriptions. They do not address the inverse problem considered here: removing non-renderable report discourse while preserving the radiographic findings specified by the source prompt.

We test this possibility with a CXR-adapted Sana generator that remains frozen throughout. Direct source prompting yields a RadDINO-FID of 54.225. Reformulating each prompt once with an unmodified Qwen3-4B reduces it by 49.2\% to 27.572 without updating either model. This result exposes a substantial optimization space outside the image generator.

\begin{figure}[t]
    \centering
    \includegraphics[width=\columnwidth]{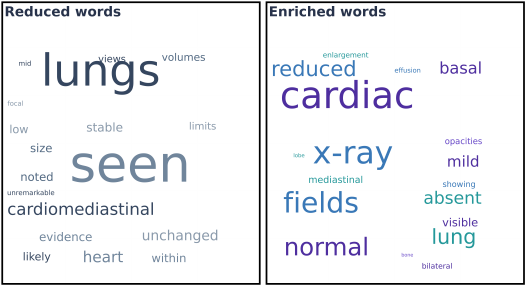}
    \caption{\textbf{Lexical shift induced by LLM rewriting.} Word clouds show the 18 words with the largest decreases and increases in the proportion of prompts containing them after rewriting. Word size reflects the magnitude of change.}
    \label{fig:prompt-lexical-cloud}
\end{figure}

\begin{figure}[t]
    \centering
    \includegraphics[width=\columnwidth]{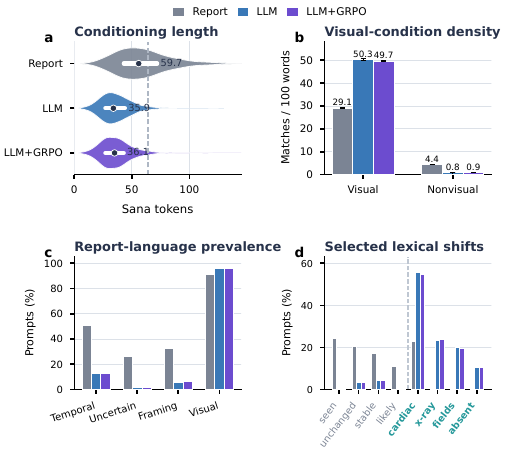}
    \caption{\textbf{Behavioral evidence of visually focused prompting.} Across 4,352 paired prompts, LLM rewriting shortens prompts (a), increases visual-term density (b), reduces temporal, uncertain, and report-framing language while preserving explicit findings (c), and favors image-description vocabulary (d). Gray, blue, and purple denote source prompting, instruction-only rewriting, and \method, respectively.}
    \label{fig:prompt-mechanism}
\end{figure}

We next examine the language change associated with this gain. Figure~\ref{fig:prompt-lexical-cloud} shows that temporal and report-oriented words become less frequent, whereas image-description terms become more prominent. Across 4,352 paired prompts, Figure~\ref{fig:prompt-mechanism} further shows that LLM reformulation shortens prompts, increases visual-term density, and suppresses temporal comparisons, uncertainty, and report framing. The resulting prompts concentrate on findings that can appear in a single radiograph. However, unconstrained reformulation reduces BioViL-T alignment with source prompts from 0.695 to 0.609, revealing a trade-off between visual focus and source-prompt alignment.

To address this trade-off, we introduce \method. A Qwen3-4B prompt policy samples candidate prompts, frozen Sana renders them under matched seeds, and radiology-aware image feedback supplies relative rewards. Standard Group Relative Policy Optimization (GRPO) updates only the prompt policy. Figure~\ref{fig:discovery} illustrates the trained policy, which restores the requested laterality under the same frozen generator. On CheXGenBench, \method reduces RadDINO-FID to 26.780 while maintaining alignment with source prompts (0.696 versus 0.695). It also achieves state-of-the-art distribution coverage and downstream classification utility. These results show that target-domain adaptation does not exhaust the optimization space and establish prompt-policy optimization as a distinct axis for text-conditioned CXR generation.

Our contributions are:
\begin{itemize}
    \item We identify an underexplored optimization dimension: LLM reformulation nearly halves RadDINO-FID for a frozen, CXR-adapted generator.
    \item Prompt analyses connect this gain to visually focused conditioning and reveal the alignment loss caused by unconstrained reformulation.
    \item We introduce \method, which applies GRPO only to the prompt policy and achieves state-of-the-art fidelity, coverage, and downstream utility while keeping Sana frozen.
\end{itemize}

\begin{figure*}[t]
    \centering
    \includegraphics[width=\textwidth]{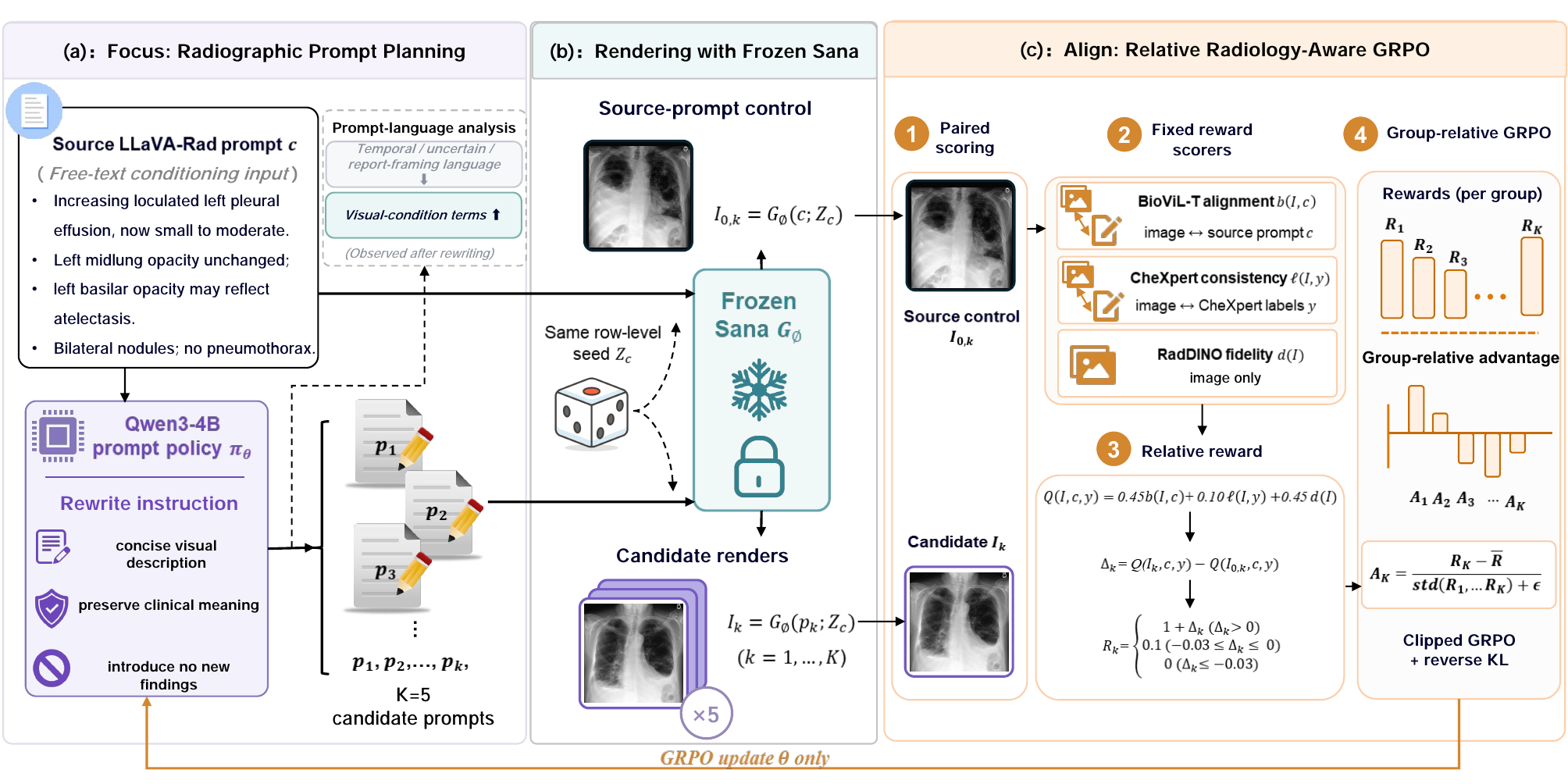}
    \caption{\textbf{JustLLMGRPO overview.} The prompt policy samples candidate prompts, and frozen Sana produces matched-seed candidate and source renderings. A radiology-aware relative reward scores each candidate, and GRPO updates only the LLM prompt planner. The optimized prompt conditions Sana at inference.}
    \label{fig:method}
\end{figure*}

\section{Related Work}

\paragraph{Chest X-ray generation.}
Text-conditioned CXR generation has advanced primarily through generator design and domain adaptation. Cheff targets high-resolution synthesis, ViewXGen adds view-specific conditioning, and RoentGen adapts a general vision--language diffusion model to radiographs~\citep{xu2024medsyn}. LLM-CXR unifies CXR understanding and generation, while RadEdit constructs counterfactual abnormalities through image editing~\citep{perez-garcia2025radedit}. CXRL further applies reinforcement learning to update the generator and its conditioning representations. CheXGenBench evaluates these systems across fidelity, alignment, coverage, privacy, and utility using radiology-aware representations~\citep{bannur2023learning,perez-garcia2025exploring}. Despite their different architectures and objectives, these approaches primarily treat the generator as the unit of adaptation. We use the CXR-adapted Sana checkpoint and CheXGenBench protocol but optimize its input prompts instead.

\paragraph{Automatic prompt optimization.}
Promptist, BeautifulPrompt, OPT2I, and VisualPrompter optimize generic prompts through rewriting, expansion, or visual correction~\citep{hao2023optimizing,cao2023beautifulprompt,manas2024improving,wu2026visualprompter,deng2022rlprompt,zhou2022large}. Our setting instead requires selective compression of detailed report-derived prompts.

\paragraph{GRPO for prompt policies.}
GRPO estimates advantages within sampled groups without a separate critic~\citep{shao2024deepseekmathc,guo2025deepseekr1}. RePrompt applies GRPO to prompt rewriting for a fixed general-purpose image generator, enriching short prompts with absolute visual rewards~\citep{wu2026reprompt}. \method differs in both the prompt regime and the learning signal. It reformulates report-derived CXR prompts and evaluates each candidate against a matched source rendering generated with the same seed. This relative radiology-aware feedback updates only the LLM prompt policy while the CXR generator remains frozen.

\begin{table*}[!t]
\centering
\scriptsize
\setlength{\tabcolsep}{2.5pt}
\begin{tabular*}{\textwidth}{@{\extracolsep{\fill}}lrrrrrrrr@{}}
\toprule
& \multicolumn{2}{c}{Fidelity} & Alignment & \multicolumn{4}{c}{PRDC} & Utility \\
\cmidrule(lr){2-3}\cmidrule(lr){4-4}\cmidrule(lr){5-8}\cmidrule(lr){9-9}
Model & FID-RD $\downarrow$ & KID-RD $\downarrow$ & BioViL-T $\uparrow$ & Prec. $\uparrow$ & Recall $\uparrow$ & Density $\uparrow$ & Coverage $\uparrow$ & Macro AUC $\uparrow$ \\
\midrule
SD V1-4 & 125.186 & 0.172 & 0.357 & 0.488 & 0.301 & 0.236 & 0.217 & 0.684 \\
SD V1-5 & 118.932 & 0.147 & 0.326 & 0.536 & 0.473 & 0.242 & 0.256 & 0.703 \\
SD V2 & 194.724 & 0.376 & 0.311 & 0.480 & 0.086 & 0.166 & 0.057 & 0.654 \\
SD V2-1 & 186.530 & 0.413 & 0.197 & 0.530 & 0.049 & 0.180 & 0.038 & 0.644 \\
RadEdit & 69.695 & 0.033 & 0.677 & 0.397 & \underline{0.544} & 0.150 & 0.285 & 0.709 \\
PixArt-$\Sigma$ & 60.154 & 0.023 & \textbf{0.697} & 0.666 & 0.522 & 0.506 & 0.506 & 0.711 \\
Sana & \underline{54.225} & \underline{0.016} & 0.695 & 0.674 & \textbf{0.614} & 0.520 & \underline{0.548} & \underline{0.730} \\
SD V3.5 Med. & 91.302 & 0.103 & 0.044 & 0.632 & 0.205 & 0.401 & 0.244 & 0.544 \\
Lumina 2.0 & 101.198 & 0.110 & 0.121 & 0.574 & 0.014 & 0.256 & 0.170 & 0.533 \\
FLUX.1-Dev & 122.400 & 0.144 & 0.036 & 0.420 & 0.008 & 0.125 & 0.326 & 0.456 \\
LLM-CXR & 71.243 & 0.061 & 0.319 & \textbf{0.782} & 0.041 & \textbf{0.671} & 0.459 & 0.676 \\
\midrule
\textbf{\method} & \textbf{26.780} & \textbf{0.014} & \underline{0.696} & \underline{0.735} & 0.498 & \underline{0.569} & \textbf{0.588} & \textbf{0.746} \\
\bottomrule
\end{tabular*}
\caption{Global fidelity, distribution coverage, and downstream utility following CheXGenBench. BioViL-T alignment is evaluated against the source prompt. CheXGenBench provides the baseline values. Bold and underline denote first- and second-ranked point estimates at the displayed precision, not statistical significance.}
\label{tab:global-fidelity}
\end{table*}

\section{Method}

Figure~\ref{fig:method} presents an overview of \method. During each rollout, the prompt policy samples a group of visually focused candidate prompts for each source prompt. Frozen Sana renders each candidate and a matched-seed source rendering. A radiology-aware relative reward measures improvement over the matched source rendering, and GRPO updates only the LLM prompt planner.

\subsection{LLM Prompt Policy}

Let $c$ denote a source LLaVA-Rad prompt, $y$ its CheXpert labels~\citep{irvin2019chexpert}, and $G_{\phi}$ the frozen Sana renderer~\citep{xie2024sana}. Direct source prompting produces $G_{\phi}(c)$. We parameterize the prompt policy with Qwen3-4B~\citep{yang2025qwen3b}. Given $c$, the LLM prompt planner samples a completion containing a reasoning trace and a final candidate prompt $p$. For compactness, $\pi_{\theta}(p\mid c)$ denotes the induced distribution over parsed candidate prompts. The instruction requests a concise image description that preserves the source meaning, introduces no new findings, and returns one final prompt inside an \texttt{<optimized\_prompt>} block. Missing or malformed completions receive zero reward. Section~\ref{sec:prompt-template} provides the complete message template.

We initialize $\pi_{\theta}$ from the unmodified Qwen3-4B model. Applying the same instruction without policy optimization defines the LLM rewriting baseline and isolates instruction-guided reformulation. In \method, candidate prompts are sampled on-policy during each rollout, and only the prompt policy is updated from image-derived rewards.

\subsection{Matched-Seed Relative Image Reward}

For each source prompt, the policy samples $K=5$ candidate prompts $p_1,\ldots,p_K$. To control stochastic variation in rendering, all candidates and the source-prompt control share a deterministic seed $z_c$, derived from the global seed and stable sample metadata:
\begin{equation}
I_k=G_{\phi}(p_k;z_c), \qquad I_{0,k}=G_{\phi}(c;z_c).
\end{equation}
Thus, all candidates for a source prompt are compared against the same matched-seed source rendering. Each generated image receives a BioViL-T alignment score $b(I,c)$, a CheXpert label-consistency score $\ell(I,y)$, and a RadDINO fidelity score $d(I)$. We define the radiology-aware image score as
\begin{equation}
Q(I,c,y)=0.45b(I,c)+0.10\ell(I,y)+0.45d(I).
\end{equation}
BioViL-T is computed against the source prompt rather than the candidate prompt. The RadDINO term combines global feature-density agreement, local reference similarity, and a nearest-neighbor memorization penalty. We define the relative improvement of candidate $k$ as $\Delta_k=Q(I_k,c,y)-Q(I_{0,k},c,y)$ and map it to
\begin{equation}
R_k=
\begin{cases}
1+\Delta_k, & \Delta_k>0,\\
0.1, & -0.03\leq\Delta_k\leq0,\\
0, & \Delta_k<-0.03.
\end{cases}
\label{eq:paired-reward}
\end{equation}
This transformation separates improvements, near-ties, and clear regressions before group-wise reward normalization.

\subsection{GRPO for Prompt Optimization}

Let $\rho$ denote the reward transformation in Equation~\ref{eq:paired-reward}. For a candidate prompt $p$, its matched-seed improvement over direct source prompting is
\begin{equation}
\begin{aligned}
\Delta Q_{\phi}(p,c,y)={}&Q(G_{\phi}(p;z_c),c,y)\\
&-Q(G_{\phi}(c;z_c),c,y).
\end{aligned}
\end{equation}
The prompt policy maximizes
\begin{equation}
\max_{\theta}\;\mathbb{E}_{(c,y)\sim\mathcal D,\,p\sim\pi_{\theta}(\cdot\mid c)}
\left[\rho\!\left(\Delta Q_{\phi}(p,c,y)\right)\right],
\qquad \phi\ \text{fixed}.
\label{eq:prompt-objective}
\end{equation}
Because $G_{\phi}$ is frozen, the renderer is treated as a black-box environment. The image-derived reward updates $\pi_{\theta}$ only through the sampled response tokens.

For each source prompt, we obtain group-normalized advantages by standardizing the five candidate rewards:
\begin{equation}
A_k=\frac{R_k-\bar R}{\operatorname{std}(R_1,\ldots,R_K)+\epsilon}.
\end{equation}
For token $w_{k,t}$, let $r_{k,t}=\exp(\log\pi_{\theta}-\log\pi_{\mathrm{old}})$ denote the importance ratio between the current and old policies. We optimize the clipped GRPO surrogate with clip ratio $0.2$ and KL regularization against the reference policy:
\begin{equation}
\begin{aligned}
\mathcal{L}_{\mathrm{prompt}}={}&-\mathbb{E}_{k,t}\left[
\min\left(r_{k,t}A_k,\right.\right.\\
&\left.\left.\operatorname{clip}(r_{k,t},0.8,1.2)A_k\right)\right]
+\beta\widehat D_{\mathrm{KL}},
\end{aligned}
\end{equation}
where $\beta=10^{-3}$ and $\widehat D_{\mathrm{KL}}$ estimates $D_{\mathrm{KL}}(\pi_{\theta}\|\pi_{\mathrm{ref}})$. The frozen reference policy $\pi_{\mathrm{ref}}$ is initialized from the same Qwen3-4B checkpoint. Source-prompt tokens are masked. All generated tokens, including the reasoning trace and final prompt, receive the sequence-level advantage. The loss is averaged over valid response tokens. Neither the reward nor the policy objective contains a prompt-length term.

\subsection{Training and Inference}

We update all parameters of Qwen3-4B-Thinking-2507 while keeping Sana frozen. Training and inference use the same message template. During training, the policy samples five completions per source prompt; at inference, it generates one completion and extracts the optimized prompt. A fully valid rollout batch contains 64 source prompts and performs 320 candidate and 320 matched source executions across 384 unique prompt--seed conditions: 320 candidate conditions and 64 source controls. Malformed completions receive zero reward and reduce the number of rendered candidate--source pairs. We perform one policy-update epoch and one optimizer step per rollout. The optimized prompt conditions the same frozen Sana renderer at inference.

\begin{table*}[!t]
\centering
\scriptsize
\setlength{\tabcolsep}{1.5pt}
\begin{tabular*}{\textwidth}{@{\extracolsep{\fill}}lrrrrrrrrrrrrrrr@{}}
\toprule
Training images & Atel. & Card. & Cons. & Edema & EC & Frac. & LL & LO & NF & PE & PO & PN & PT & SD & Avg. \\
\midrule
Real & 0.750 & 0.760 & 0.720 & 0.850 & 0.610 & 0.580 & 0.630 & 0.700 & 0.840 & 0.840 & 0.740 & 0.670 & 0.710 & 0.830 & 0.731 \\
\midrule
SD V1-4 & 0.700 & 0.700 & 0.670 & 0.810 & 0.560 & 0.570 & 0.630 & 0.670 & 0.800 & 0.770 & 0.650 & 0.600 & 0.650 & 0.800 & 0.684 \\
SD V1-5 & 0.720 & 0.720 & 0.690 & 0.810 & 0.600 & 0.530 & \underline{0.660} & 0.670 & 0.820 & 0.790 & 0.680 & 0.620 & 0.700 & \underline{0.830} & 0.703 \\
SD V2 & 0.660 & 0.690 & 0.660 & 0.780 & \underline{0.610} & 0.530 & 0.550 & 0.630 & 0.750 & 0.760 & 0.500 & 0.610 & 0.640 & 0.780 & 0.654 \\
SD V2-1 & 0.630 & 0.670 & 0.650 & 0.710 & 0.550 & 0.590 & 0.620 & 0.620 & 0.750 & 0.740 & 0.570 & 0.560 & 0.610 & 0.750 & 0.644 \\
RadEdit & 0.730 & \underline{0.730} & \underline{0.720} & 0.840 & \underline{0.610} & 0.560 & 0.600 & \underline{0.690} & 0.810 & \underline{0.820} & \underline{0.720} & \underline{0.660} & 0.660 & 0.770 & 0.709 \\
PixArt-$\Sigma$ & \underline{0.740} & \underline{0.730} & 0.700 & 0.840 & \underline{0.610} & 0.580 & 0.610 & \underline{0.690} & \underline{0.830} & 0.810 & 0.680 & 0.630 & 0.700 & 0.800 & 0.711 \\
Sana & \underline{0.740} & \textbf{0.760} & \underline{0.720} & \underline{0.850} & \underline{0.610} & \underline{0.620} & 0.630 & \textbf{0.700} & \underline{0.830} & \textbf{0.840} & \textbf{0.730} & 0.640 & \underline{0.720} & \underline{0.830} & \underline{0.730} \\
SD V3.5 & 0.550 & 0.550 & 0.560 & 0.550 & 0.470 & 0.470 & 0.470 & 0.530 & 0.600 & 0.540 & 0.580 & 0.490 & 0.550 & 0.710 & 0.544 \\
Lumina 2.0 & 0.460 & 0.480 & 0.520 & 0.510 & 0.460 & 0.570 & 0.530 & 0.520 & 0.590 & 0.550 & 0.570 & 0.490 & 0.500 & 0.710 & 0.533 \\
FLUX.1-Dev & 0.410 & 0.410 & 0.440 & 0.400 & 0.440 & 0.520 & 0.480 & 0.420 & 0.400 & 0.380 & 0.500 & 0.480 & 0.440 & 0.670 & 0.456 \\
LLM-CXR & 0.700 & 0.690 & 0.700 & 0.810 & \underline{0.610} & 0.570 & 0.540 & 0.650 & 0.800 & 0.770 & 0.660 & 0.610 & 0.630 & 0.730 & 0.676 \\
\midrule
\textbf{\method} & \textbf{0.750} & \textbf{0.760} & \textbf{0.740} & \textbf{0.860} & \textbf{0.650} & \textbf{0.630} & \textbf{0.680} & \textbf{0.700} & \textbf{0.840} & \textbf{0.840} & \textbf{0.730} & \textbf{0.680} & \textbf{0.740} & \textbf{0.850} & \textbf{0.746} \\
\bottomrule
\end{tabular*}
\caption{Per-condition downstream classification AUC after training ResNet-50 on 20,000 synthetic CXRs and evaluating on held-out real CXRs. \method is evaluated on the development set; CheXGenBench provides the baselines and real-data reference. Bold and underline denote first- and second-ranked synthetic point estimates at the displayed precision, with ties retained; they do not indicate significance.}
\label{tab:downstream-auc}
\end{table*}

\section{Experiments}

\subsection{Experimental Setup}

\paragraph{Data and evaluation protocol.}
We follow the MIMIC-CXR split and LLaVA-Rad annotations released with CheXGenBench~\citep{johnson2019mimiccxr,dutt2026chexgenbench,zambranochaves2025clinically}. The training set contains 237,388 samples, and the evaluation set contains 5,034 samples corresponding to 4,352 unique source prompts.

\paragraph{Models and optimization.}
The image generator is the CheXGenBench Sana checkpoint, adapted to CXR synthesis for 20 epochs and then frozen. We generate $512\times512$ images with 20 denoising steps and guidance scale 4.5. We train the Qwen3-4B prompt policy with AdamW at a learning rate of $10^{-6}$ and a maximum completion length of 2,048 tokens. Each rollout performs one policy-update epoch and one optimizer step, using temperature 1, top-$p=1$, and no top-$k$ truncation. Two GPUs host the tensor-parallel policy, and three workers perform generation and reward evaluation. Sana and all reward models remain frozen throughout optimization.

\paragraph{Baselines and metrics.}
We compare against the eleven CheXGenBench generators: four Stable Diffusion variants, RadEdit, PixArt-$\Sigma$, Sana, SD3.5 Medium, Lumina 2.0, FLUX.1-Dev, and LLM-CXR. CheXGenBench provides the baseline and real-data values. The Sana baseline conditions the same frozen generator directly on source prompts. Metrics include RadDINO-based FID and KID, BioViL-T alignment with source prompts, PRDC, conditional RadDINO-FID, and downstream classification AUC. All evaluations use the metric implementations released with CheXGenBench.

\subsection{Benchmark Results}

\paragraph{Global CheXGenBench results.}
On this development evaluation, Table~\ref{tab:global-fidelity} shows that \method achieves state-of-the-art performance among the compared CheXGenBench systems, with the best point estimates for RadDINO-FID (26.780), KID-RD (0.014), PRDC coverage (0.588), and macro AUC (0.746). It also ranks second in BioViL-T alignment, precision, and density. Relative to direct Sana prompting, RadDINO-FID decreases by 50.6\% while alignment with source prompts remains comparable (0.696 versus 0.695). PRDC recall decreases from 0.614 to 0.498, so the improvement is not uniform across distributional metrics; aggregate BioViL-T alignment also does not establish finding-level correctness.

\paragraph{Downstream utility.}
Following CheXGenBench, we train a ResNet-50 on each 20,000-image synthetic dataset and test it on held-out real CXRs. \method is best or tied among synthetic generators for all fourteen conditions at two-decimal precision (Table~\ref{tab:downstream-auc}). Its macro AUC is 0.746, compared with 0.730 for Sana and 0.731 for the real-data reference. The synthetic images therefore support downstream pathology classification, although these point estimates do not establish superiority over real-data training.

\paragraph{Condition-level fidelity.}
Global RadDINO-FID can obscure pathology-specific behavior. Table~\ref{tab:conditional-fid} therefore reports conditional RadDINO-FID for fourteen CheXpert categories. \method achieves the best score for Cardiomegaly, Consolidation, Enlarged Cardiomediastinum, No Finding, and Pneumonia, and ranks second for five additional categories. It falls outside the top two for Fracture, Lung Lesion, Pleural Other, and Pneumothorax. Compared with direct Sana, it improves five categories and worsens nine; competitive rankings therefore do not imply uniform improvement.

\begin{table*}[!t]
\centering
\scriptsize
\setlength{\tabcolsep}{1.65pt}
\begin{tabular*}{\textwidth}{@{\extracolsep{\fill}}lrrrrrrrrrrrrrr@{}}
\toprule
Model & Atel. & Card. & Cons. & Edema & EC & Frac. & LL & LO & NF & PE & PO & PN & PT & SD \\
\midrule
SD V1-4 & 134.11 & 131.04 & 184.30 & 144.84 & 217.75 & 238.78 & 225.99 & 129.38 & 106.34 & 128.16 & 255.84 & 163.82 & 212.48 & 135.10 \\
SD V1-5 & 125.67 & 124.75 & 181.25 & 139.94 & 213.94 & 243.17 & \textbf{123.13} & 255.64 & 167.81 & 193.75 & 243.17 & 101.08 & \textbf{119.91} & 123.64 \\
SD V2 & 188.72 & 193.91 & 241.24 & 214.40 & 214.40 & 253.91 & 268.28 & 280.11 & 193.99 & 299.48 & 223.43 & 250.96 & 183.34 & 193.99 \\
SD V2-1 & 179.20 & 181.79 & 228.43 & 193.62 & 242.65 & 263.01 & 260.15 & 185.00 & 192.30 & 178.84 & 287.27 & 213.26 & 242.60 & 176.99 \\
RadEdit & 63.38 & 62.79 & 136.59 & 76.94 & 155.97 & 197.58 & 184.11 & 61.90 & 67.88 & 60.60 & 215.92 & 114.66 & 151.34 & 53.10 \\
PixArt-$\Sigma$ & 59.27 & 60.39 & 133.96 & 73.93 & 155.53 & 179.44 & 174.63 & 56.83 & 48.74 & 59.05 & 210.90 & 108.42 & 150.55 & 51.61 \\
Sana & \textbf{51.03} & \underline{54.68} & \underline{127.46} & \textbf{67.84} & \underline{147.00} & \underline{172.32} & \underline{163.14} & \textbf{49.23} & \underline{44.60} & \textbf{49.80} & \textbf{199.45} & \underline{99.52} & \underline{141.99} & \textbf{46.51} \\
SD V3.5 Med. & 94.94 & 94.84 & 149.05 & 111.94 & 168.48 & 184.75 & 173.37 & 86.72 & 89.60 & 91.92 & 203.62 & 124.07 & 163.27 & 86.99 \\
Lumina 2.0 & 109.39 & 111.11 & 162.36 & 131.18 & 182.35 & 191.83 & 182.22 & 99.53 & 95.66 & 105.25 & 213.50 & 134.58 & 165.09 & 102.78 \\
FLUX.1-Dev & 137.10 & 133.60 & 176.76 & 152.91 & 191.48 & 191.02 & 194.97 & 133.37 & 100.58 & 137.66 & 221.23 & 156.59 & 190.93 & 127.03 \\
LLM-CXR & 71.57 & 71.37 & 136.65 & 83.18 & 148.28 & \textbf{168.50} & 163.22 & 66.93 & 64.62 & 67.83 & \underline{200.84} & 108.04 & 147.52 & 67.54 \\
\midrule
\textbf{\method} & \underline{52.42} & \textbf{54.14} & \textbf{126.88} & \underline{68.48} & \textbf{145.65} & 176.32 & 164.66 & \underline{50.75} & \textbf{43.70} & \underline{51.94} & 204.80 & \textbf{99.26} & 142.82 & \underline{46.93} \\
\bottomrule
\end{tabular*}
\caption{Conditional RadDINO-FID ($\downarrow$) across fourteen potentially overlapping CheXpert categories. \method is evaluated on the development set. Bold and underline denote first- and second-ranked point estimates at the displayed precision, not significance. EC: Enlarged Cardiomediastinum; LL: Lung Lesion; LO: Lung Opacity; NF: No Finding; PE: Pleural Effusion; PO: Pleural Other; PN: Pneumonia; PT: Pneumothorax; SD: Support Devices.}
\label{tab:conditional-fid}
\end{table*}

\subsection{Prompt Reformulation and GRPO}

Table~\ref{tab:stage-ablation} compares source prompting, instruction-only reformulation, and GRPO under the same frozen Sana generator. Instruction-only rewriting reduces RadDINO-FID by 49.2\%, from 54.225 to 27.572, but decreases BioViL-T alignment with source prompts by 0.086. GRPO further reduces RadDINO-FID by 2.9\% and restores aggregate alignment to 0.696. Thus, most fidelity improvement is already present after rewriting, while policy optimization recovers the associated alignment loss without updating Sana.

\subsection{Prompt-Language Analysis}

To characterize the language change associated with the large instruction-only fidelity gain, we compare 4,352 unique LLaVA-Rad source prompts, deterministic rewrites from the unmodified Qwen3-4B planner, and prompts from the GRPO-trained policy. The instruction-only baseline uses the same message template with greedy decoding, and token counts are computed with Sana's Gemma tokenizer. The rewrite instruction permits compression, but neither the reward nor the policy objective contains a prompt-length term.

Mean length decreases from 59.7 source-prompt tokens to 35.9 after instruction-only rewriting and remains 36.1 after GRPO (Figure~\ref{fig:prompt-mechanism}a). The fraction of prompts longer than 64 tokens falls from 35.9\% to 3.0\% and 3.3\%, respectively. Compression is systematic: 98.5\% of instruction-only prompts and 96.4\% of \method prompts are shorter than their source prompts. No prompt exceeds Sana's 300-token limit; the observed shift therefore does not arise from hard truncation.

Compression is accompanied by a change in prompt composition. We define a visual-conditioning lexicon comprising visible findings, anatomy, spatial expressions, devices, and negations, and a non-visual-discourse lexicon comprising temporal comparisons, uncertainty, and report framing. Visual-conditioning term density increases from 29.1 to 50.3 and 49.7 terms per 100 words across source prompting, instruction-only rewriting, and \method, while non-visual-discourse term density decreases from 4.4 to 0.8 and 0.9 (Figure~\ref{fig:prompt-mechanism}b).

The category analysis identifies which expressions are reduced. The prevalence of temporal or comparison language falls from 50.6\% to 12.6\% and 12.9\%; uncertainty falls from 26.7\% to 1.7\% and 2.0\%; and report framing falls from 32.3\% to 6.0\% and 6.3\%. Conversely, the fraction of prompts containing at least one explicit visual-finding term increases from 90.9\% to 96.2\% and 95.9\% (Figure~\ref{fig:prompt-mechanism}c). This statistic measures the prevalence of visual language, not whether every finding in the source prompt is preserved.

\begin{table}[tbp]
\centering
\small
\setlength{\tabcolsep}{2.6pt}
\begin{tabular*}{\columnwidth}{@{\extracolsep{\fill}}lccrr@{}}
\toprule
System & Planner & GRPO & FID-RD $\downarrow$ & Align. $\uparrow$ \\
\midrule
LLaVA-Rad & -- & No & 54.225 & \underline{0.695} \\
LLM rewriting & Qwen3-4B & No & \underline{27.572} & 0.609 \\
\textbf{\method} & Qwen3-4B & Yes & \textbf{26.780} & \textbf{0.696} \\
\bottomrule
\end{tabular*}
\caption{Effect of instruction-guided prompt reformulation and subsequent GRPO under the same frozen Sana generator. Alignment is evaluated against the source prompt. Bold and underline denote first- and second-ranked point estimates; they do not indicate significance.}
\label{tab:stage-ablation}
\end{table}

Figure~\ref{fig:prompt-lexical-cloud} shows the corresponding lexical reorganization. Temporal and report-oriented words such as \emph{seen}, \emph{unchanged}, \emph{stable}, and \emph{likely} become less prevalent. Anatomical phrasing shifts from forms such as \emph{heart}, \emph{lungs}, and \emph{cardiomediastinal} toward \emph{cardiac}, \emph{lung}, and \emph{mediastinal}, while image-description terms such as \emph{x-ray}, \emph{fields}, \emph{normal}, and \emph{absent} become more frequent. The word clouds summarize changes in usage and do not by themselves establish that clinical concepts were deleted or introduced.

Across both the quantitative panels and lexical summary, instruction-only and GRPO-optimized prompts remain similar. Most observed compression and lexical change is therefore already present under instruction-only rewriting. GRPO neither shortens prompts further nor increases measured visual-conditioning term density, yet BioViL-T alignment with source prompts rises from 0.609 to 0.696. This result suggests that the benefit of policy optimization may depend on finer changes in which findings and qualifiers are expressed or realized in the generated image, which are not captured by prompt length or the fixed lexicons.

\subsection{Qualitative Analysis}

\begin{figure}[tbp]
    \centering
    \includegraphics[width=\columnwidth]{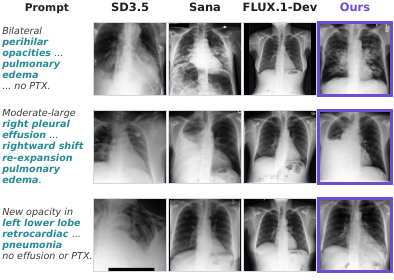}
    \caption{\textbf{Representative qualitative comparisons under matched source prompts.} Each row starts from the same LLaVA-Rad source prompt; cyan italics mark the requested findings. \method more clearly depicts bilateral perihilar opacities, a right pleural effusion, and a left retrocardiac pneumonia opacity than the displayed baselines.}
    \label{fig:qualitative-cases}
\end{figure}

Figure~\ref{fig:qualitative-cases} presents three representative cases. Each row starts from the same source prompt: SD3.5, Sana, and FLUX.1-Dev use it directly, whereas \method uses its optimized rewrite. The panel illustrates the capability of the final system rather than the incremental effect of GRPO; the supplement provides 98 source-prompt-matched cases.

Across the three cases, \method most clearly realizes the requested radiographic pattern: bilateral perihilar opacities consistent with pulmonary edema, a moderate-to-large right pleural effusion with associated re-expansion opacity, and a focal left lower-lobe retrocardiac opacity suggesting pneumonia. The displayed baselines either weaken the abnormality, alter its distribution, or produce conspicuous artifacts. In all three examples, the optimized rewrite retains the principal finding, laterality, and relevant negative constraints from the source prompt.

\subsection{Prompt-Policy versus Generator Optimization}

To examine the optimization locus, we compare with a direct Sana-GRPO control that updates rank-16 Sana LoRA modules for 950--1,000 outer steps using 144 image samples per optimization prompt. It uses the same reward components and weights, but an absolute objective, different computational budget, and different trainable parameterization. Table~\ref{tab:optimization-locus} is therefore descriptive rather than a compute-matched ablation.

Direct Sana-GRPO raises BioViL-T alignment to 0.816, but RadDINO-FID rises to 59.145 and PRDC recall and coverage fall to 0.080 and 0.384. \method instead obtains 26.780, 0.498, and 0.588, respectively. In this control, higher reward-model alignment therefore coincides with lower fidelity and distributional coverage, without establishing a general advantage over compute-matched generator optimization.

\begin{table}[tbp]
\centering
\scriptsize
\setlength{\tabcolsep}{2.3pt}
\begin{tabular*}{\columnwidth}{@{\extracolsep{\fill}}llrrrr@{}}
\toprule
System & Update & FID-RD $\downarrow$ & Align. $\uparrow$ & Recall $\uparrow$ & Cov. $\uparrow$ \\
\midrule
LLaVA-Rad & -- & \underline{54.225} & 0.695 & \textbf{0.614} & \underline{0.548} \\
Sana-GRPO & Sana LoRA & 59.145 & \textbf{0.816} & 0.080 & 0.384 \\
\textbf{\method} & Qwen3-4B & \textbf{26.780} & \underline{0.696} & \underline{0.498} & \textbf{0.588} \\
\bottomrule
\end{tabular*}
\caption{Prompt-policy and direct-generator optimization using the same reward components and weights but different objectives, computational budgets, and trainable parameterizations. Sana-GRPO uses the composite reward. Alignment is BioViL-T against the source prompt. Bold and underline denote ranked point estimates, not significance.}
\label{tab:optimization-locus}
\end{table}

\subsection{Further Analysis}

\paragraph{Preservation of source information.}
An automatic audit finds 90.3\% laterality preservation and 97.8\% preservation of negative-label polarity. Together with the recovery of source-prompt alignment from 0.609 to 0.696, these results indicate that \method preserves generation-critical constraints while focusing the frozen generator on renderable findings (Table~\ref{tab:semantic-audit}). The audit is sensitive to surface-form changes and does not measure clinical correctness.

\paragraph{Reward sensitivity and extended qualitative evidence.}
The direct Sana-GRPO sweep likewise shows that BioViL-T and composite rewards raise alignment to 0.827 and 0.816 while recall remains 0.104 and 0.080 and coverage 0.413 and 0.384 (Table~\ref{tab:direct-sana-reward-sweep}). Section~\ref{app:qualitative} provides 98 source-prompt-matched cases across all fourteen categories. Together, these analyses separate fidelity, preservation of source information, and distributional coverage. Sections~\ref{app:prompt-policy} and~\ref{app:reward-protocol} provide additional prompt-policy and data details.

\subsection{Prompt Template}
\label{sec:prompt-template}

Training and inference share this Qwen chat template. The source prompt is inserted at \texttt{\{original\_prompt\}}.

\begin{tcolorbox}[
    enhanced,
    breakable,
    title={Shared Qwen prompt},
    colback=PromptBackground,
    colframe=PromptPurple,
    colbacktitle=PromptPurple,
    coltitle=white,
    fonttitle=\bfseries\footnotesize,
    boxrule=0.55pt,
    arc=1.2mm,
    left=6pt,
    right=6pt,
    top=5pt,
    bottom=5pt
]
\footnotesize\raggedright\color{PromptInk}
{\bfseries\textcolor{PromptPurple}{System message}}\par
{\ttfamily\textcolor{PromptBlue}{You are an expert prompt optimizer for chest X-ray diffusion models.}}\par
\medskip
{\bfseries\textcolor{PromptBlue}{User message}}\par
{\ttfamily
\textcolor{PromptBlue}{Source diffusion prompt:}\par
\textcolor{PromptCyan}{\bfseries\{original\_prompt\}}\par
\smallskip
Rewrite the source prompt into a chest X-ray prompt that is better suited for diffusion generation.\par
\smallskip
\textcolor{PromptCyan}{\bfseries Guidelines:}\par
\textcolor{PromptCyan}{--} \textbf{Preserve} the source meaning.\par
\textcolor{PromptCyan}{--} Feel free to rewrite, compress, and reorder the wording to make it better suited for chest X-ray generation.\par
\textcolor{PromptCyan}{--} \textbf{Do not add} new clinical facts that are not already present in the source.\par
\smallskip
You can reason in \textcolor{PromptBlue}{<think></think>}.\par
After \textcolor{PromptBlue}{</think>}, put the final result in \textbf{exactly one} \textcolor{PromptPurple}{<optimized\_prompt></optimized\_prompt>} block.\par
\textcolor{PromptPurple}{\bfseries Do not output any other text.}}
\end{tcolorbox}

\section{Conclusion}

This work identifies prompt formulation as an underexplored source of performance in text-conditioned CXR generation. With a CXR-adapted Sana generator frozen, instruction-guided LLM reformulation nearly halves RadDINO-FID, showing that domain adaptation leaves substantial optimization space. Prompt analyses link this gain to visible radiographic findings and reveal alignment loss under unconstrained reformulation.

Building on this finding, \method uses standard GRPO and group-relative radiology-aware image feedback to optimize only the prompt policy. On CheXGenBench, it reduces RadDINO-FID from 54.225 to 26.780, maintains source-prompt alignment, and achieves state-of-the-art distribution coverage and downstream classification utility. These results establish prompt-policy optimization as a distinct axis, showing that prompt expression can determine a substantial part of the achievable quality.

\bibliography{aaai2027}

\clearpage
\appendix
\flushbottom
\setcounter{secnumdepth}{2}
\setcounter{dbltopnumber}{4}
\numberwithin{figure}{section}
\numberwithin{table}{section}

\newcommand{\qualpage}[3]{%
\begin{figure*}[p]
\centering
\includegraphics[width=\textwidth]{#1}
\caption{Matched qualitative comparisons for \textbf{#2}. In each row, the baselines use the source prompt directly, whereas \method uses the corresponding optimized prompt. Purple frames denote \method.}
\label{fig:#3}
\end{figure*}
}

\section*{Appendix}
\subsection*{Overview}

This appendix provides reproducibility details and additional analyses. Section~\ref{app:prompt-policy} documents prompt construction, inference, and optimization details for the prompt policy. Section~\ref{app:reward-protocol} specifies the matched-seed reward and data protocol. Section~\ref{app:further-analysis} reports an automatic source-prompt preservation audit and the complete direct Sana reward sweep. Section~\ref{app:qualitative} presents 98 matched qualitative comparisons across all 14 CheXpert categories.

\section{Prompt Policy Configuration and Inference}
\label{app:prompt-policy}

\subsection{Prompt Construction and Inference}

\paragraph{Shared message template.}
Training and inference use the same Qwen chat template, with the source prompt provided as the user input. CheXpert labels, generated images, and reward values are excluded from the policy input and are introduced only after sampling to score the generated images.

\paragraph{Training-time output handling.}
During training, the prompt-policy output is parsed to extract the candidate prompt. Empty or invalid outputs receive zero reward and are not rendered. Valid candidate prompts are normalized and limited to 1,200 characters before being passed to Sana.

The parser first removes generation-control tokens and joins nonempty lines, then searches for a single closing \texttt{</think>} marker followed by a nonempty \texttt{<optimized\_prompt>} block. Whitespace is normalized without changing the lexical content of the candidate. Completions that contain multiple candidate blocks, an empty block, or no closing reasoning marker are treated as malformed. This conservative rule prevents partial reasoning traces or formatting artifacts from entering the image generator. The same parser is used during validation and benchmark inference, so the reported fallback behavior is not specific to training.

\paragraph{Benchmark inference.}
At benchmark inference, the optimized prompt policy greedily generates one candidate prompt for each unique source prompt. The candidate prompt conditions Sana. If no valid candidate is produced, the system falls back to the source prompt. Each unique source prompt is rewritten once, and the resulting candidate prompt is reused for all evaluation samples associated with that source prompt.

Inference therefore has two deterministic stages: prompt rewriting and image rendering. The rewrite stage is deduplicated by the source prompt string, whereas the rendering stage preserves the evaluation-row identifier when deriving the seed. This distinction avoids repeatedly sampling the language policy for duplicate rows while retaining row-level reproducibility for the generated images. In the released implementation, greedy decoding is used for benchmark inference; stochastic decoding is reserved for on-policy training rollouts.

\subsection{Optimization Configuration}

Table~\ref{tab:optimization-config} summarizes the configuration of the Qwen3-4B-Thinking-2507 prompt policy. The trainable policy is initialized from the pretrained model, and all Qwen parameters are updated. A frozen copy of the initial model serves as the reference policy for KL regularization. Sana and all reward models remain frozen throughout optimization.

\begin{table*}[t]
\centering
\caption{Prompt-policy optimization and rendering configuration.}
\label{tab:optimization-config}
\small
\setlength{\tabcolsep}{7pt}
\renewcommand{\arraystretch}{1.16}
\begin{tabular}{@{}L{0.27\textwidth}L{0.68\textwidth}@{}}
\toprule
Setting & Value \\
\midrule
Prompt policy & Qwen3-4B-Thinking-2507; full-parameter optimization \\
Source prompts per rollout batch & 64 \\
Candidate prompts per source prompt & 5 \\
Optimizer & AdamW, learning rate $10^{-6}$ \\
GRPO update & One update epoch and one optimizer step per rollout batch \\
GRPO objective & Clip ratio $0.2$; KL coefficient $10^{-3}$ against the frozen reference policy \\
Maximum input/output lengths & 768 input tokens; 2,048 output tokens \\
Rollout decoding & Temperature $1.0$; top-$p=1.0$; no top-$k$ truncation \\
Inference decoding & Greedy decoding; maximum 2,048 new tokens \\
Sana renderer & CXR-adapted Sana model fine-tuned for 20 epochs and then frozen \\
Image generation & $512\times512$; 20 denoising steps; guidance scale $4.5$ \\
Compute allocation & Two GPUs for tensor-parallel policy rollout and optimization; three workers for Sana rendering and reward evaluation \\
\bottomrule
\end{tabular}
\end{table*}

\section{Matched-Seed Reward and Data Protocol}
\label{app:reward-protocol}

\subsection{Deterministic Matched-Seed Comparison}

For each candidate prompt, frozen Sana renders a candidate image and a source-prompt control from the same initial noise. A deterministic seed derived from the source prompt and evaluation-sample identifier is shared by each pair. This matched-seed design holds stochastic initialization fixed when measuring the effect of prompt reformulation. The five candidates sampled for each source prompt share the same source-control rendering, avoiding redundant control generation. Candidates that do not yield a valid prompt are omitted.

\subsection{Radiology-Aware Relative Reward}

For an image $I$, source prompt $c$, and CheXpert labels $y$, the radiology-aware score combines BioViL-T alignment, CheXpert label consistency, and RadDINO fidelity with weights $0.45$, $0.10$, and $0.45$, respectively. BioViL-T is computed against the source prompt rather than the candidate prompt. In the label-consistency term, positive labels contribute the predicted probability, negative labels contribute one minus that probability, and uncertain labels are ignored.

The RadDINO component combines global feature-density agreement, local top-$k$ reference similarity, and a nearest-neighbor memorization penalty with internal weights $0.60$, $0.30$, and $0.10$. Candidate improvement is computed relative to the matched source rendering and mapped to the piecewise reward in Equation~\ref{eq:paired-reward}. The five rewards within each source-prompt group are standardized to obtain group-relative advantages. Neither the reward nor the policy objective includes a prompt-length term.

The relative construction is important for interpreting the reward. A candidate is not rewarded simply for obtaining a high absolute image score; it is rewarded for improving on the source-prompt control generated with the same initial noise. This reduces sensitivity to prompt-independent variation in the frozen renderer and makes the learning signal specific to the language intervention. The source-prompt alignment term is computed against the original source prompt for both candidate and control images. Consequently, the policy is encouraged to remove report discourse that is difficult to render while retaining findings that remain supported by the source condition.

The three reward components serve complementary roles. BioViL-T supplies a continuous image--text similarity signal, the label term protects explicit positive and negative CheXpert findings, and RadDINO favors the radiographic feature distribution used by the benchmark. The label term is intentionally low-weight: it acts as a safeguard against polarity errors but does not dominate the image-level fidelity signal. The memorization penalty is applied only within the RadDINO component and is not a textual similarity constraint.

\subsection{Training and Evaluation Data}

Table~\ref{tab:data-protocol} summarizes the data used for prompt-policy training, model selection, and evaluation. The pathology-balanced validation subset supports category-aware model selection, while the complete evaluation pool supports the benchmark and prompt-language analyses.

The training pool contains 237,388 rows from the CheXGenBench MIMIC-CXR split. For model selection, the validation subset contains 560 rows balanced across the 14 CheXpert categories, while the final evaluation pool contains 5,034 rows and 4,352 unique source prompts. Prompt-language statistics are computed on the unique-prompt view of this pool, whereas image metrics are computed on evaluation rows. The downstream utility experiment uses a separate 20,000-image synthetic training subset and held-out real CXRs, following the CheXGenBench protocol. These different sample counts should not be interpreted as independent test sets: they support distinct diagnostics of prompt behavior, image quality, and utility.

\paragraph{Row construction and required fields.}
Each prompt-policy row contains the source LLaVA-Rad prompt, a stable row identifier, and the 14 CheXpert labels used by the image reward. Optional view and orientation metadata are retained for diagnostics but are not inserted into the policy message unless already expressed in the source prompt. Rows with empty source prompts are excluded during conversion to the training parquet. The CheXpert payload distinguishes positive, negative, and uncertain labels; uncertain entries are ignored by the label-consistency reward rather than being converted to either polarity.

\paragraph{Separation of policy input and reward metadata.}
Only the source prompt is visible to the LLM prompt policy. Labels, image paths, real images, reference embeddings, and reward values remain in the rollout metadata and are accessed after a completion has been parsed. This separation prevents the policy from copying structured labels or reward diagnostics into the generated prompt. It also preserves the intended deployment setting, in which inference requires only a source prompt and does not assume access to a paired real image or CheXpert annotation.

\paragraph{Duplicate prompts and evaluation units.}
The evaluation CSV contains repeated source prompts associated with different rows. The language policy rewrites each of the 4,352 unique source prompts once, after which the parsed candidate is joined back to all 5,034 evaluation rows. Image generation and scoring remain row-level because the deterministic seed also uses the row identifier. Thus, duplicate text does not trigger redundant language-model inference, but duplicate rows do not collapse into a single generated image. Prompt statistics use the unique-prompt table to avoid over-weighting duplicates, whereas FID, alignment, PRDC, and downstream analyses follow the benchmark row structure.

\paragraph{Length handling and deterministic conversion.}
CSV rows are converted in a stable order without shuffling. Chat-formatted inputs are limited to 768 tokens, with overlong examples filtered rather than truncated. Rollouts allow 2,048 output tokens, but the parser retains only the normalized \texttt{<optimized\_prompt>} content, capped at 1,200 characters before rendering. The global seed is 42, and rendering seeds are derived from the source prompt and row identifier, reproducing prompt groups and matched candidate--control comparisons.

\paragraph{Reference assets.}
The frozen CheXpert classifier and RadDINO reference cache are constructed from the designated real-CXR training subset and are not updated during prompt-policy optimization. The RadDINO cache stores normalized reference features used for global density and local-neighbor comparisons. The downstream ResNet-50 is trained separately on synthetic images and is not part of the GRPO reward. These assets serve different purposes and are kept disjoint in the implementation: reward models provide the online learning signal, while the downstream classifier evaluates the utility of a completed synthetic dataset. Reward scoring remains fixed throughout optimization.

\begin{table*}[t]
\centering
\caption{Training, model-selection, and evaluation protocol.}
\label{tab:data-protocol}
\small
\setlength{\tabcolsep}{7pt}
\renewcommand{\arraystretch}{1.16}
\begin{tabular}{@{}L{0.27\textwidth}L{0.68\textwidth}@{}}
\toprule
Component & Protocol \\
\midrule
Prompt-policy training set & 237,388 LLaVA-Rad rows from the CheXGenBench MIMIC-CXR training split \\
Validation subset & 560 pathology-balanced rows, 40 per CheXpert category \\
Evaluation set & 5,034 rows containing 4,352 unique source prompts \\
Prompt-language analysis & The same 4,352 unique prompts, compared across source prompting, instruction-only rewriting, and \method \\
Downstream utility & ResNet-50 trained on 20,000 synthetic CXRs and evaluated on held-out real CXRs following CheXGenBench \\
Statistical scope & Prompt-policy results from one training run \\
\bottomrule
\end{tabular}
\end{table*}

\FloatBarrier
\section{Further Analysis}
\label{app:further-analysis}

\subsection{Preservation of Generation-Critical Source Information}
\label{app:semantic-audit}

We apply a conservative automatic audit to all 5,034 rewritten evaluation rows. Rule-based extractors compare condition polarity, laterality, view, device, and anatomical terms between source and candidate prompts. A CheXbert-based comparison provides a complementary label-level diagnostic. Table~\ref{tab:semantic-audit} reports the aggregate preservation rates.

\begin{table}[t]
\centering
\caption{Automatic source-prompt preservation diagnostics over 5,034 rewritten evaluation rows. Positive-label preservation excludes the aggregate \emph{No Finding} label.}
\label{tab:semantic-audit}
\footnotesize
\setlength{\tabcolsep}{3pt}
\renewcommand{\arraystretch}{1.10}
\begin{tabular}{@{}L{0.72\columnwidth}r@{}}
\toprule
Diagnostic & Rate (\%) \\
\midrule
Source-mentioned condition polarity preserved & 64.1 \\
Positive CheXpert label preserved in the candidate prompt & 60.0 \\
Negative CheXpert label not flipped to positive & 97.8 \\
Laterality terms preserved & 90.3 \\
Device terms preserved & 74.1 \\
Anatomical-region terms preserved & 61.2 \\
View terms preserved & 60.6 \\
Exact CheXbert label match between source and candidate prompts & 29.9 \\
\bottomrule
\end{tabular}
\end{table}

Despite substantial prompt reformulation, \method preserves negative-label polarity in 97.8\% of cases and laterality in 90.3\%. These attributes are generation-critical constraints because they determine whether a finding should appear and where it should be rendered. Together with the recovery of source-prompt BioViL-T alignment from 0.609 under instruction-only rewriting to 0.696 after GRPO, these results show that visual focusing and source alignment are not inherently conflicting. By retaining source-grounded visual constraints while suppressing non-renderable report language, the optimized prompt provides a more actionable condition that enables frozen Sana to better reflect the source prompt.

Positive findings, devices, and anatomical regions have surface-form preservation rates of 60.0\%, 74.1\%, and 61.2\%, respectively. These values reflect the policy's compression of report-style expressions into shorter radiographic descriptions. The conservative extractor records exact listed terms, so synonym substitution and lexical consolidation appear as surface-form changes even when the visual concept remains represented. Together with the high preservation of polarity and laterality, the pattern shows that the policy prioritizes constraints that directly determine the requested image content.

The audit and image metrics provide complementary views of this transformation. The text audit tracks which source attributes remain explicit after rewriting, while BioViL-T and RadDINO evaluate how the rewritten condition affects the rendered image. Their joint pattern supports selective visual focusing: the policy preserves decisive source-grounded constraints, removes temporal and report-framing language, and expresses the remaining findings in a form that frozen Sana can use more effectively. This interpretation is consistent with the simultaneous recovery of source-prompt alignment and improvement in fidelity and coverage.

\subsection{Complete Direct Sana Reward Sweep}
\label{app:direct-sana-reward-sweep}

Table~\ref{tab:direct-sana-reward-sweep} provides the complete comparison between prompt-policy and direct-generator optimization. Each direct Sana-GRPO baseline updates rank-16 Sana LoRA modules for 950--1,000 optimization iterations and evaluates 144 generated images per optimization prompt. The sweep places generator-side and prompt-side optimization in the same metric space, making it possible to compare how different reward choices affect alignment, fidelity, recall, coverage, and downstream utility.

\begin{table*}[t]
\centering
\caption{Direct Sana-GRPO reward sweep and frozen-Sana prompt-policy baselines. FID-RD is RadDINO-FID; Recall and Cov. are PRDC recall and coverage; AUC is downstream macro AUC; Align. is source-prompt BioViL-T similarity. Dashes indicate unavailable evaluations.}
\label{tab:direct-sana-reward-sweep}
\footnotesize
\setlength{\tabcolsep}{4.5pt}
\renewcommand{\arraystretch}{1.12}
\begin{tabular*}{\textwidth}{@{\extracolsep{\fill}}lllrrrrr@{}}
\toprule
System & Updated module & Reward & FID-RD $\downarrow$ & Align. $\uparrow$ & Recall $\uparrow$ & Cov. $\uparrow$ & AUC $\uparrow$ \\
\midrule
Source prompting & None & None & 54.225 & 0.695 & 0.614 & 0.548 & 0.730 \\
Direct Sana-GRPO & Sana LoRA & PickScore & 111.748 & 0.560 & 0.001 & 0.098 & 0.633 \\
Direct Sana-GRPO & Sana LoRA & CLIPScore & 117.754 & 0.531 & 0.001 & 0.100 & 0.657 \\
Direct Sana-GRPO & Sana LoRA & HPSv2 & 138.462 & 0.476 & 0.001 & 0.040 & 0.622 \\
Direct Sana-GRPO & Sana LoRA & ImageReward & 80.508 & 0.557 & 0.009 & 0.168 & 0.670 \\
Direct Sana-GRPO & Sana LoRA & BioViL-T & 50.196 & 0.827 & 0.104 & 0.413 & 0.728 \\
Direct Sana-GRPO & Sana LoRA & Composite & 59.145 & 0.816 & 0.080 & 0.384 & 0.710 \\
LLM rewriting (no GRPO) & None & None & 27.572 & 0.609 & -- & -- & -- \\
\textbf{\method} & Qwen3-4B & Composite & \textbf{26.780} & 0.696 & 0.498 & \textbf{0.588} & \textbf{0.746} \\
\bottomrule
\end{tabular*}
\end{table*}

General-purpose preference rewards yield FID-RD values from 80.508 to 138.462, while BioViL-T and composite rewards raise alignment to 0.827 and 0.816. \method achieves an FID-RD of 26.780, recall of 0.498, coverage of 0.588, and source-prompt alignment of 0.696. The sweep shows that applying radiographic alignment, label consistency, and feature-distribution feedback to the prompt policy produces the strongest joint fidelity, coverage, and downstream utility in the table. The gain is associated with learning how to express radiographic content for the frozen generator.

\section{Additional Qualitative Comparisons}
\label{app:qualitative}

We report 98 matched comparisons, with seven cases for each of the 14 CheXpert categories. Each row contains an abridged source prompt, the paired real MIMIC-CXR image, five baseline generations conditioned on the source prompt, and the \method output from frozen Sana. Teal text marks category-relevant terms, and purple frames identify \method. Rows are matched by source prompt across all displayed systems, making polarity, laterality, and finding prominence directly comparable.

The gallery is drawn from completed evaluation outputs using an automatic ranking based on category response, source-prompt alignment, image realism, quality checks, and margins over the displayed baselines. Every page follows the same order: source prompt, real CXR, SD1.5, PixArt-$\Sigma$, SD3.5, Sana, FLUX.1-Dev, and \method. Prompts, seeds, model order, and generated images remain fixed, preserving a direct comparison of conditioning strategies.

Across pleural, parenchymal, cardiomediastinal, focal-lesion, fracture, normal-examination, and support-device categories, \method strengthens category-relevant visual evidence while retaining source-specified polarity and laterality. The gallery complements the aggregate results by showing that the learned prompt policy improves the conditioning signal across heterogeneous radiographic targets while the Sana generator and sampling configuration remain unchanged.

\paragraph{Prompt-to-image mechanism.}
The matched layout makes the effect of prompt reformulation visible at the image level. The source-prompt columns show how several generators respond to the original report-derived condition, while the final column isolates the effect of replacing that condition with the \method rewrite under frozen Sana. Because the generator weights, sampling configuration, and stored outputs are fixed, changes in the prominence, location, and extent of a requested finding can be traced to the optimized language condition. This visual evidence complements the lexical analysis in the main paper: shorter prompts with denser radiographic terminology provide a more direct interface between clinical text and the frozen renderer.

\paragraph{Coverage of radiographic attributes.}
The 14 category pages jointly cover positive and negative findings, unilateral and bilateral abnormalities, focal and diffuse patterns, global cardiomediastinal morphology, and device-related structure. Pleural effusion and pneumothorax emphasize polarity and laterality; edema, consolidation, pneumonia, atelectasis, and lung opacity emphasize spatial distribution; cardiomegaly and enlarged cardiomediastinum emphasize global silhouette; and fracture, lung lesion, and support devices emphasize localized structure. The \emph{No Finding} page provides the complementary normal-examination setting. Together, these comparisons show that the prompt policy operates across different levels of radiographic granularity rather than relying on a category-specific phrasing template.

\paragraph{Matched comparison protocol.}
Within each row, the source prompt and displayed outputs remain aligned across systems, so the same requested radiographic content is inspected throughout the panel. Category highlighting identifies the words that define the principal visual target, while the fixed column order supports rapid comparison of finding presence, location, extent, and overall image structure. The real CXR supplies anatomical context, and the generated columns show how each conditioning strategy translates the same source information into an image.

\paragraph{Relation to aggregate results.}
The gallery provides an image-level view of the trends summarized by RadDINO-FID, BioViL-T alignment, PRDC coverage, and downstream utility. Improved fidelity appears as more coherent radiographic structure, maintained alignment appears as retention of the requested finding, and broader coverage appears through diverse realizations across categories and anatomical presentations. The qualitative pages therefore connect the aggregate gains to visible prompt-conditioned behavior and show how optimized radiographic wording supports the frozen generator across the complete benchmark taxonomy.

\paragraph{Consistent visual encoding.}
The same typography, category highlighting, image scale, and column labels are used on every page. Teal source terms identify the requested radiographic evidence, and the purple frame provides an immediate visual anchor for the optimized-prompt result. This shared encoding supports rapid cross-category inspection without changing the interpretation of individual columns. It also makes the appendix a direct extension of the quantitative analysis: readers can move from a benchmark metric to the corresponding category pages and inspect how the learned conditioning policy expresses that radiographic target.

\qualpage{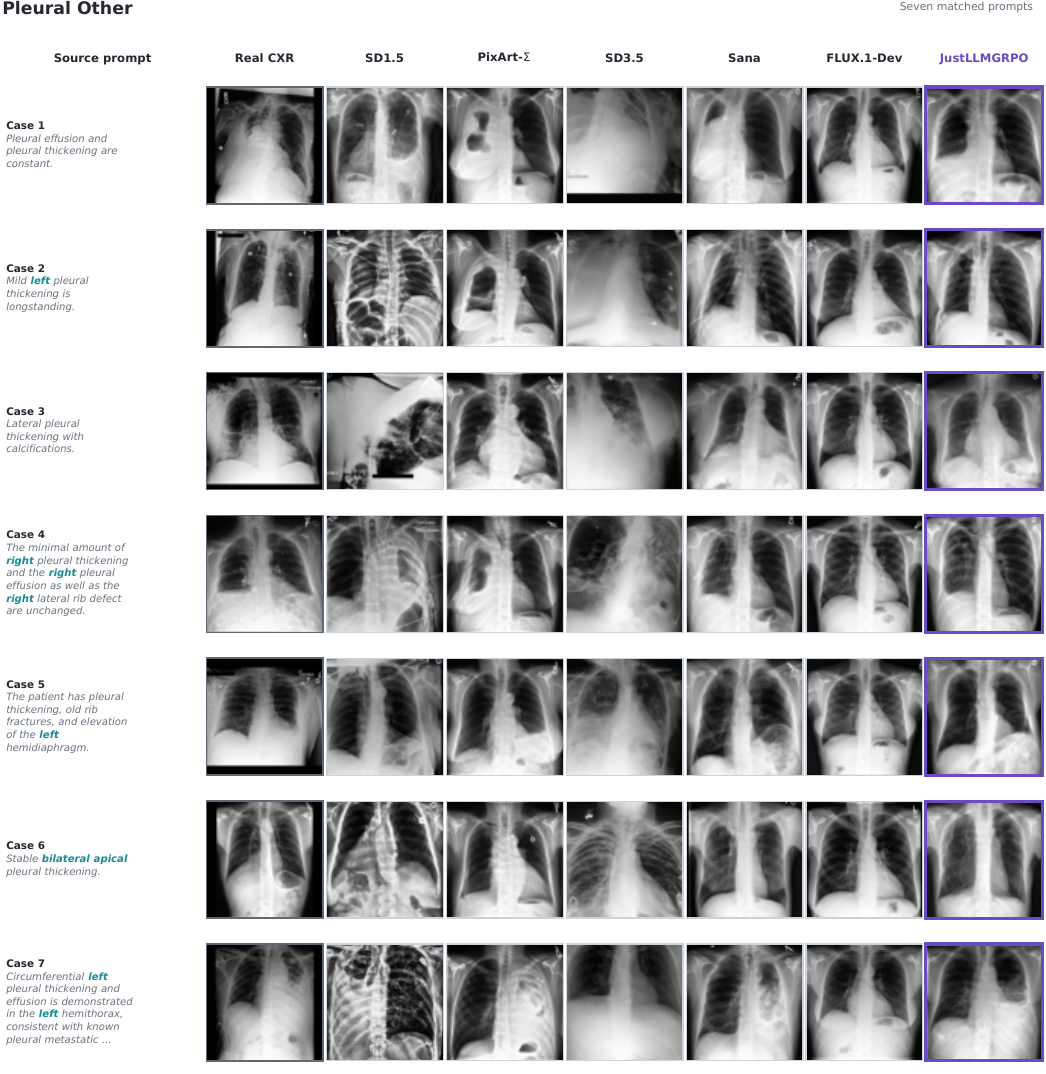}{Pleural Other}{supp-pleural-other}
\qualpage{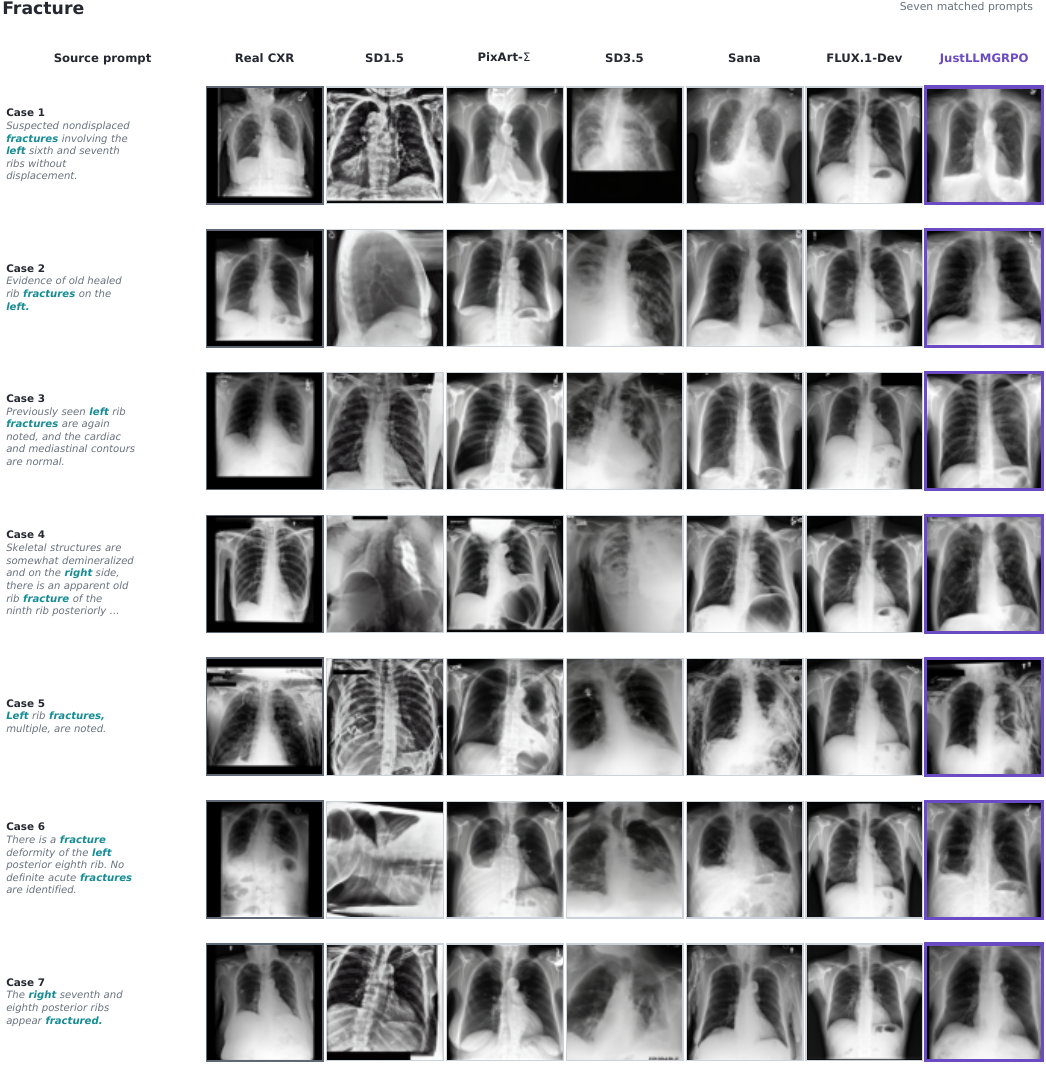}{Fracture}{supp-fracture}
\qualpage{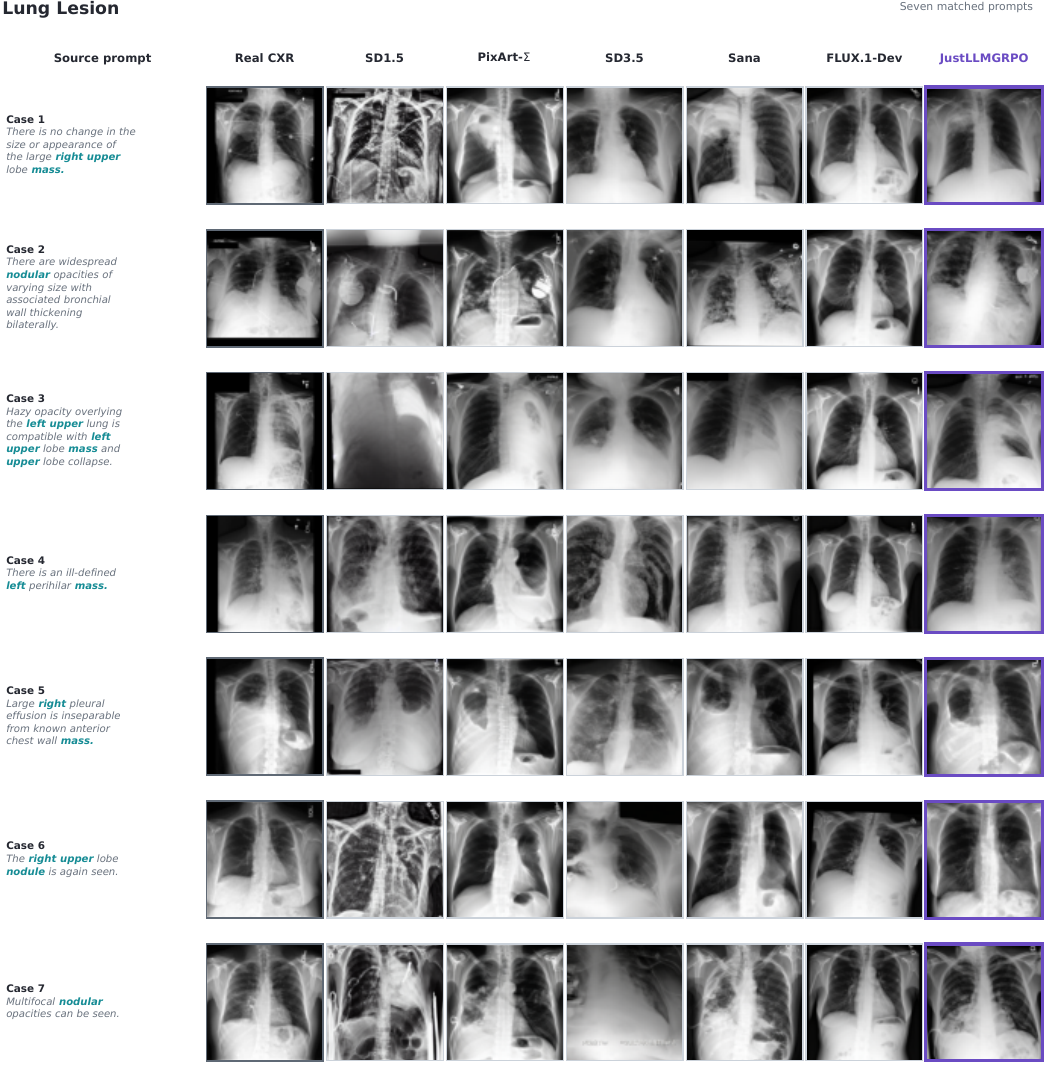}{Lung Lesion}{supp-lung-lesion}
\qualpage{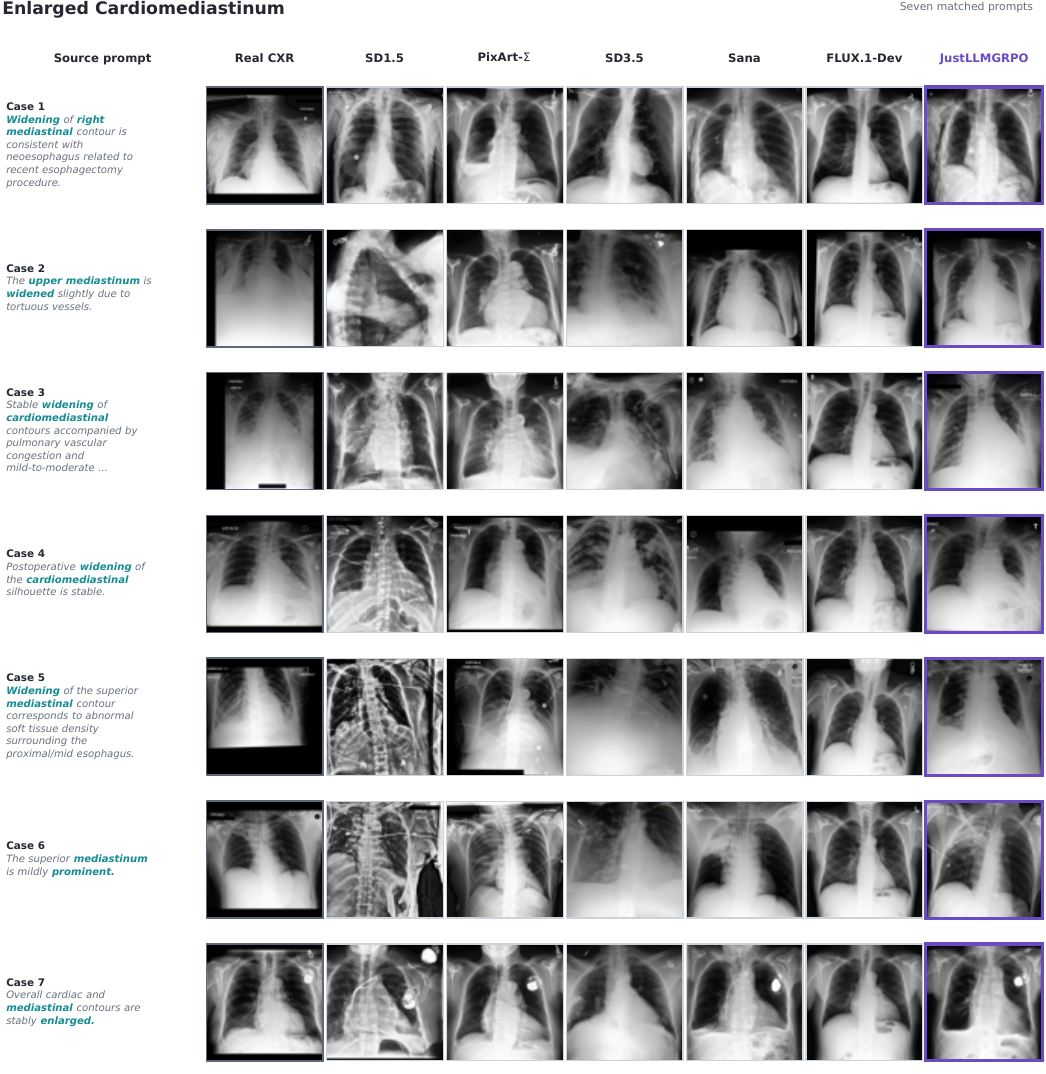}{Enlarged Cardiomediastinum}{supp-enlarged-cardiomediastinum}
\qualpage{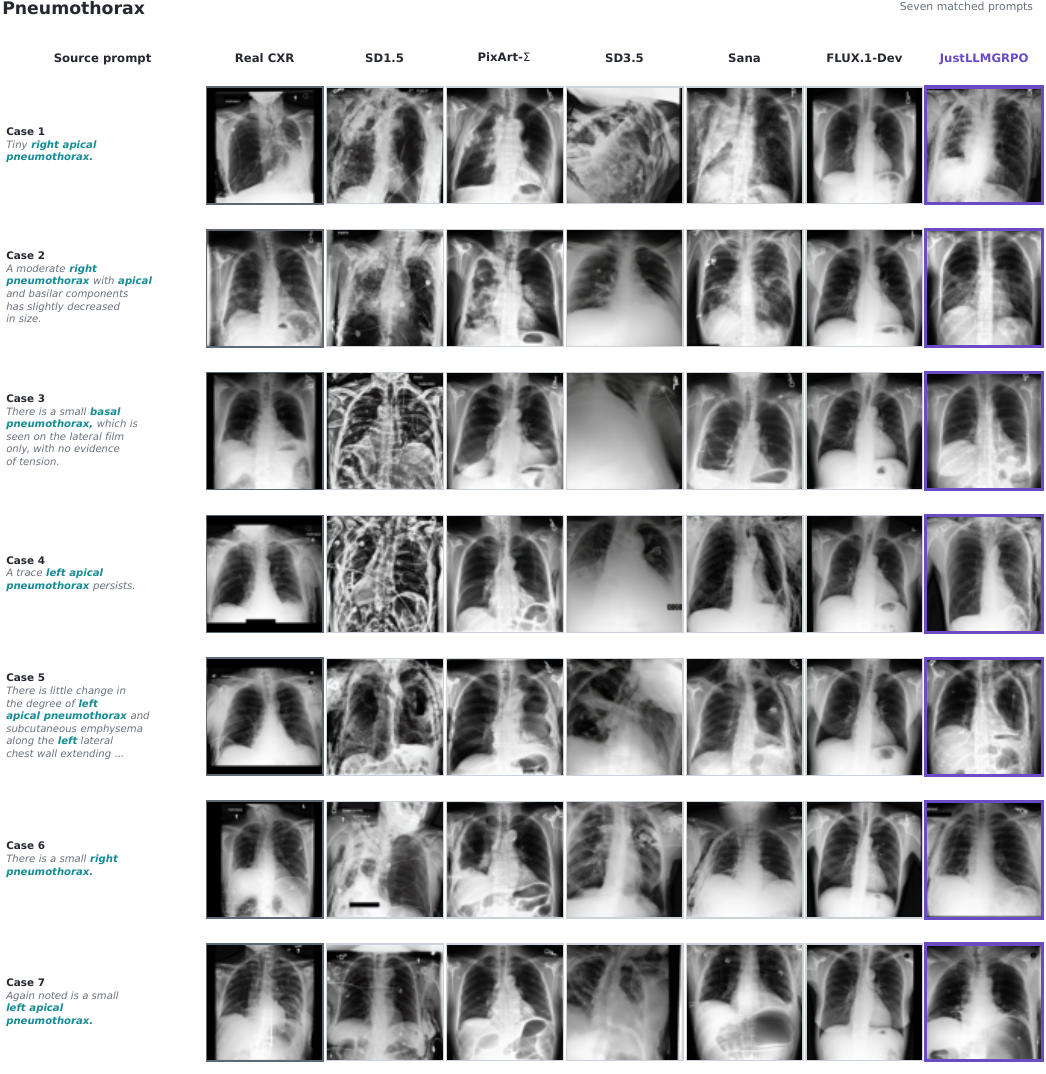}{Pneumothorax}{supp-pneumothorax}
\qualpage{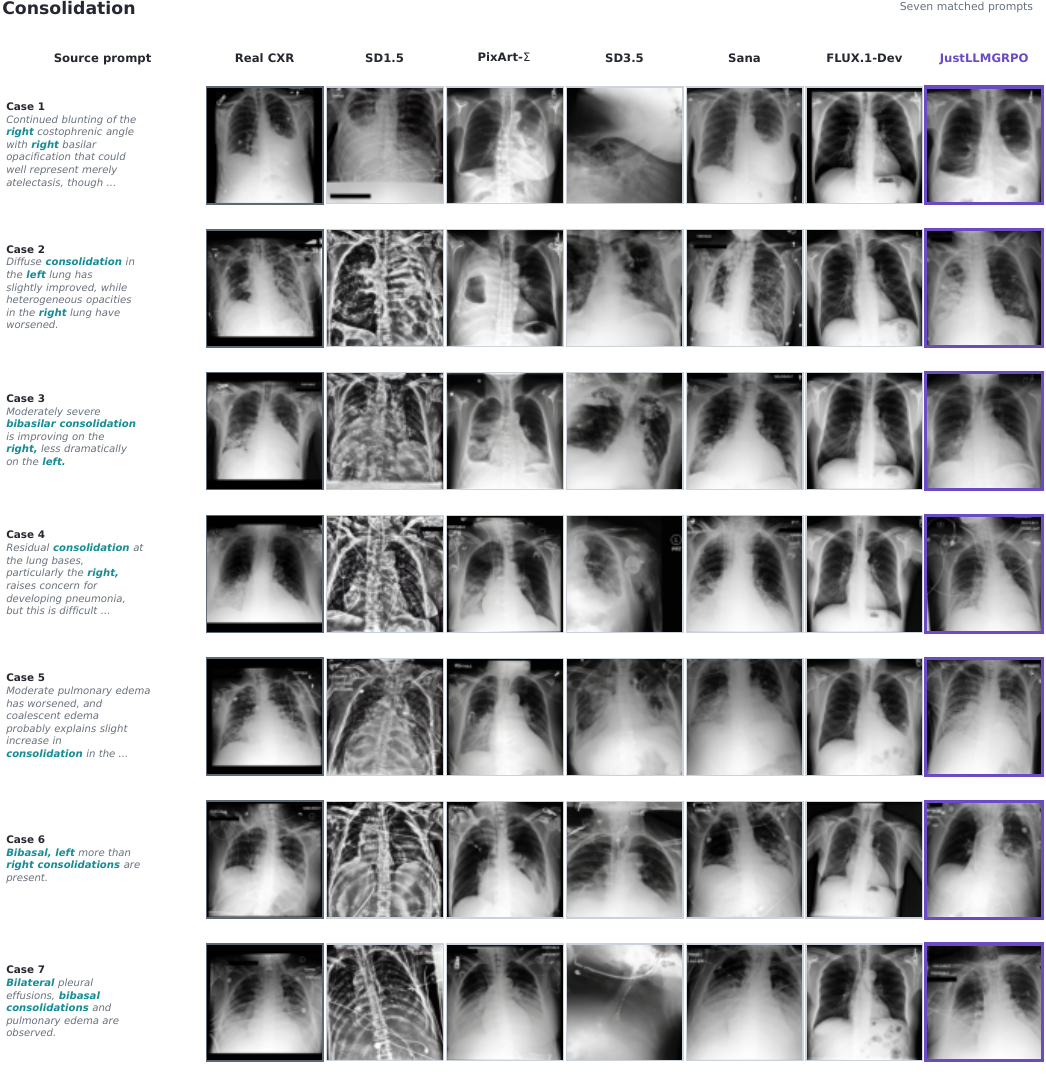}{Consolidation}{supp-consolidation}
\qualpage{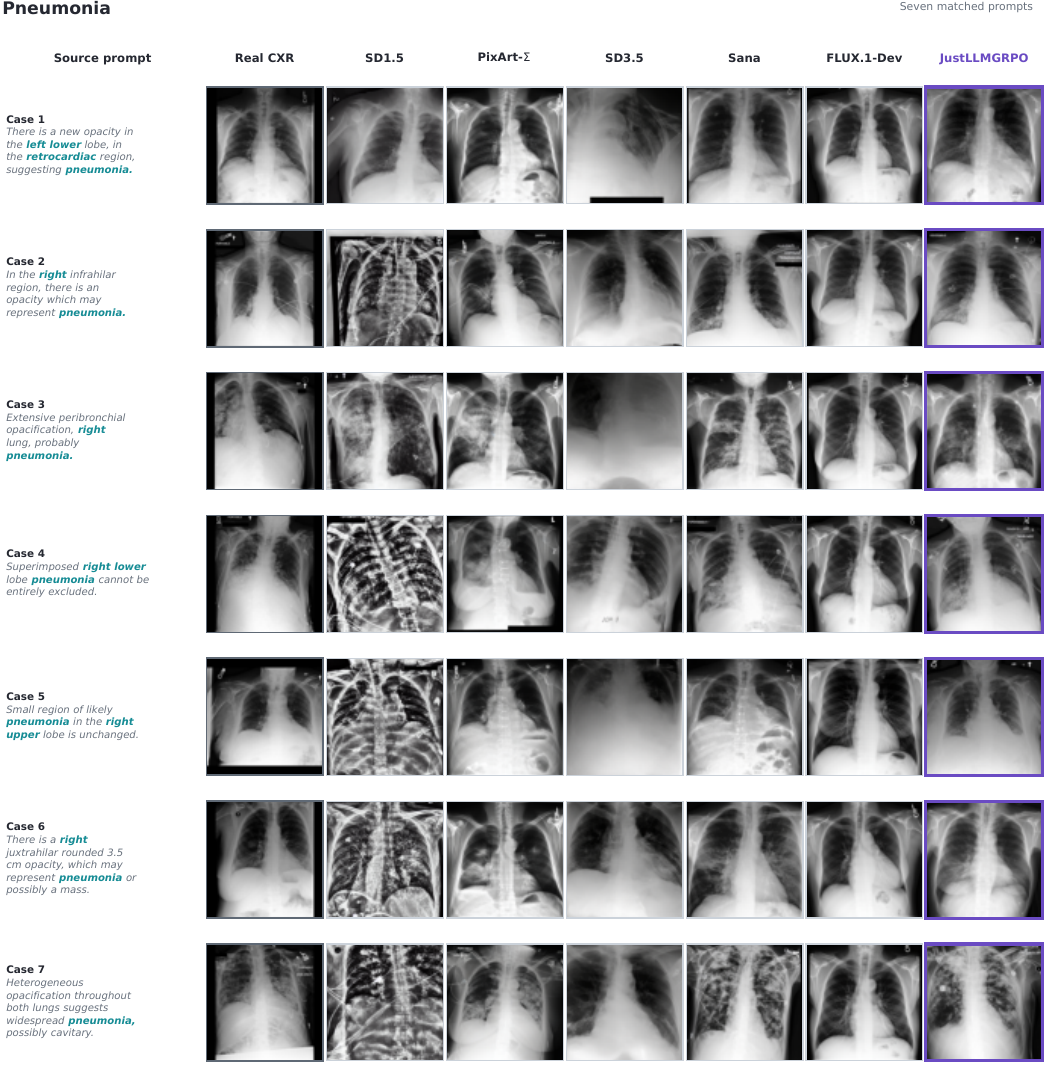}{Pneumonia}{supp-pneumonia}
\qualpage{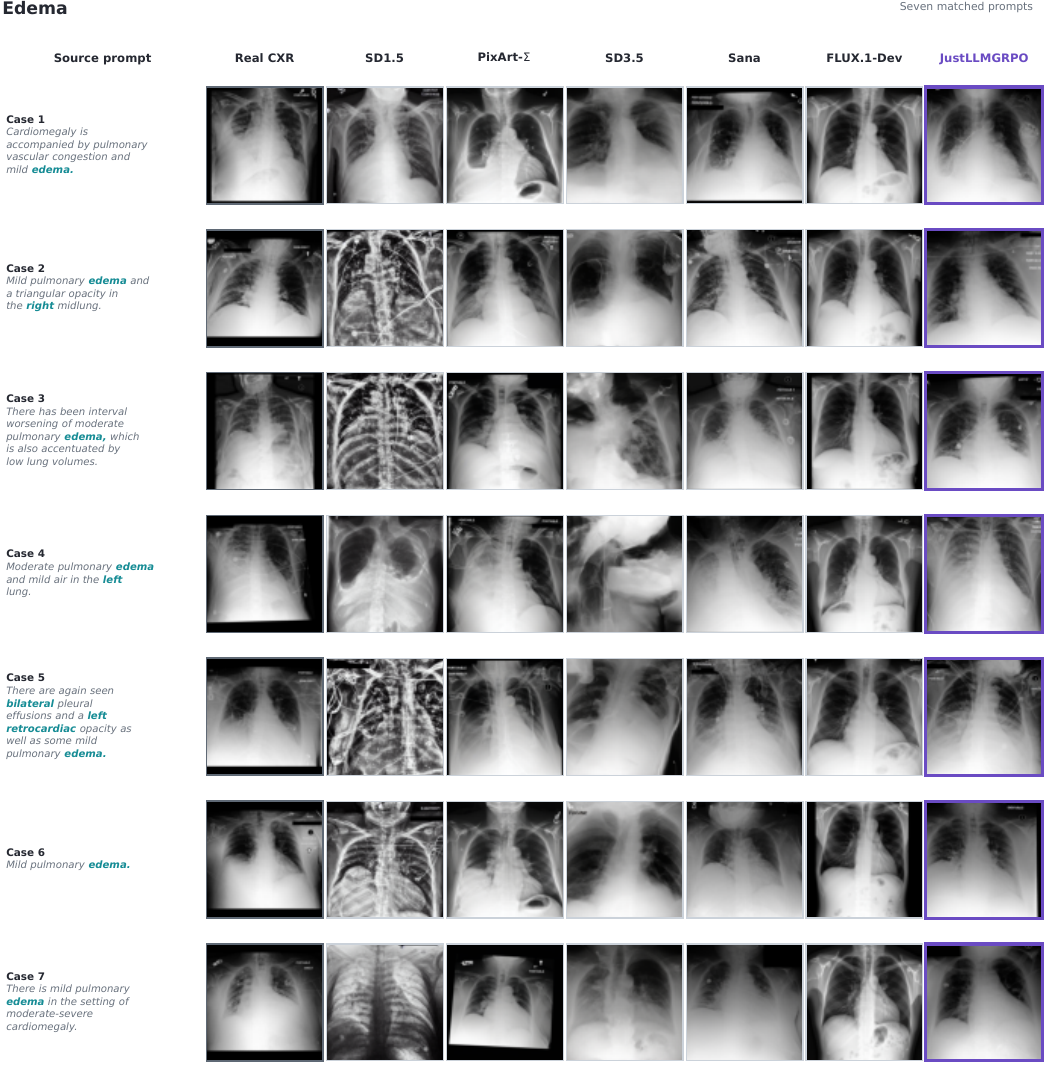}{Edema}{supp-edema}
\qualpage{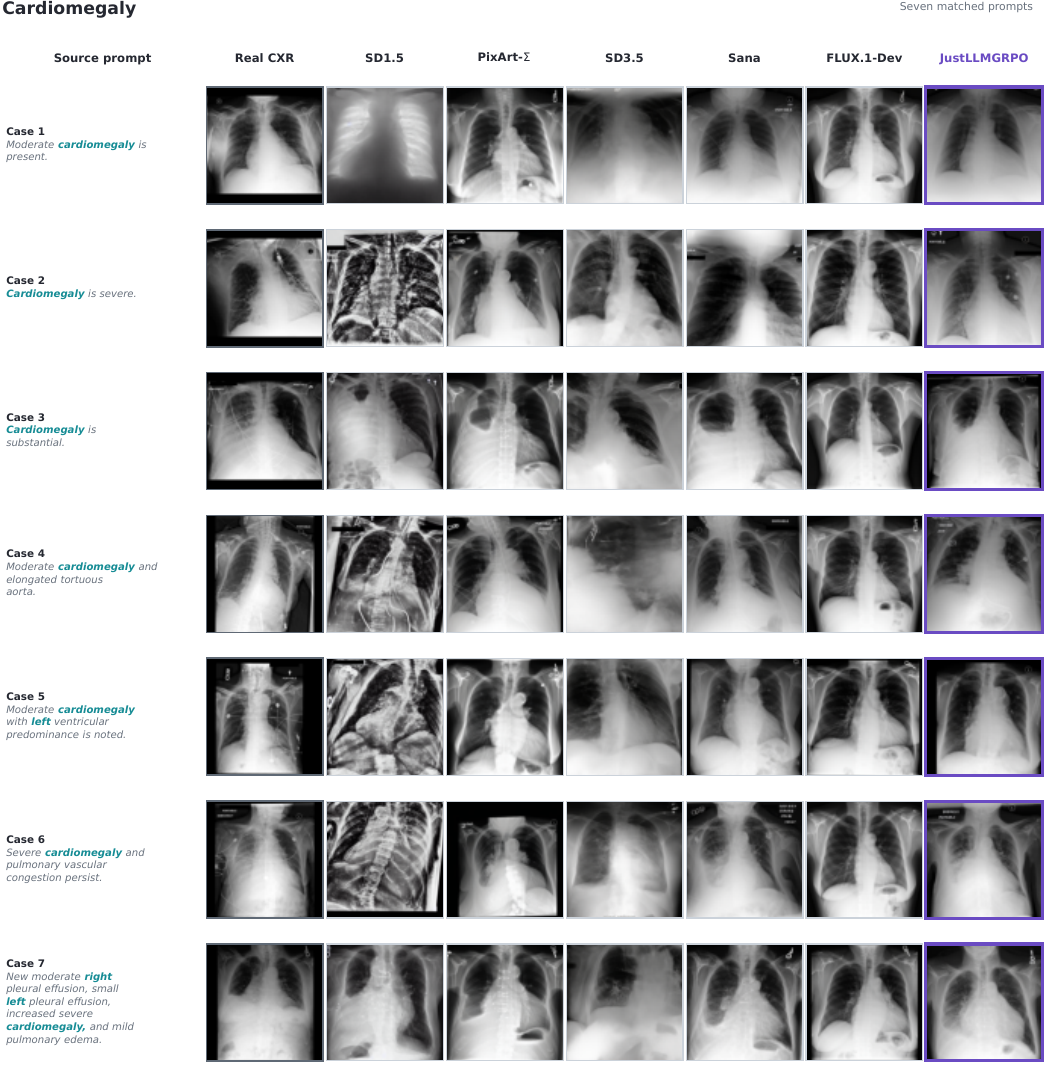}{Cardiomegaly}{supp-cardiomegaly}
\qualpage{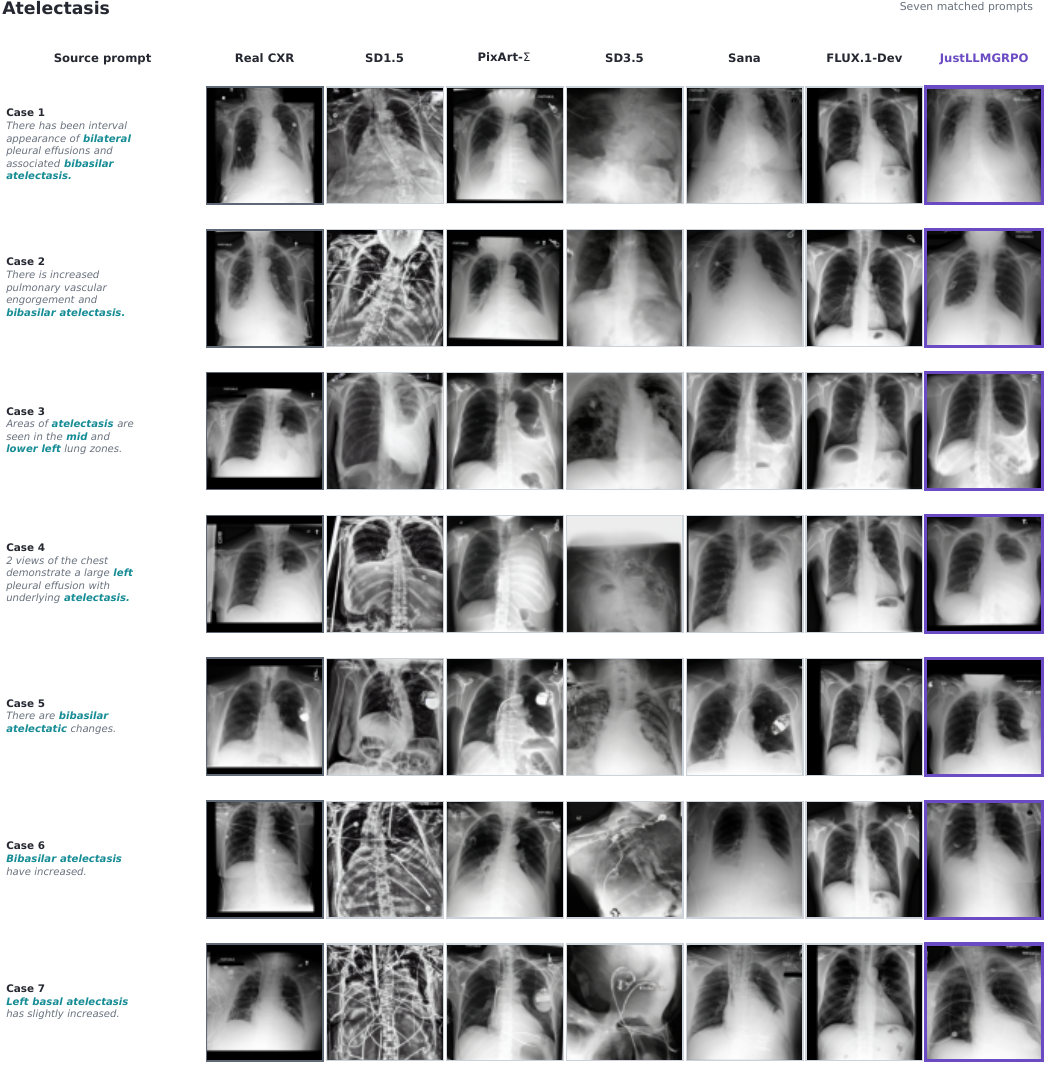}{Atelectasis}{supp-atelectasis}
\qualpage{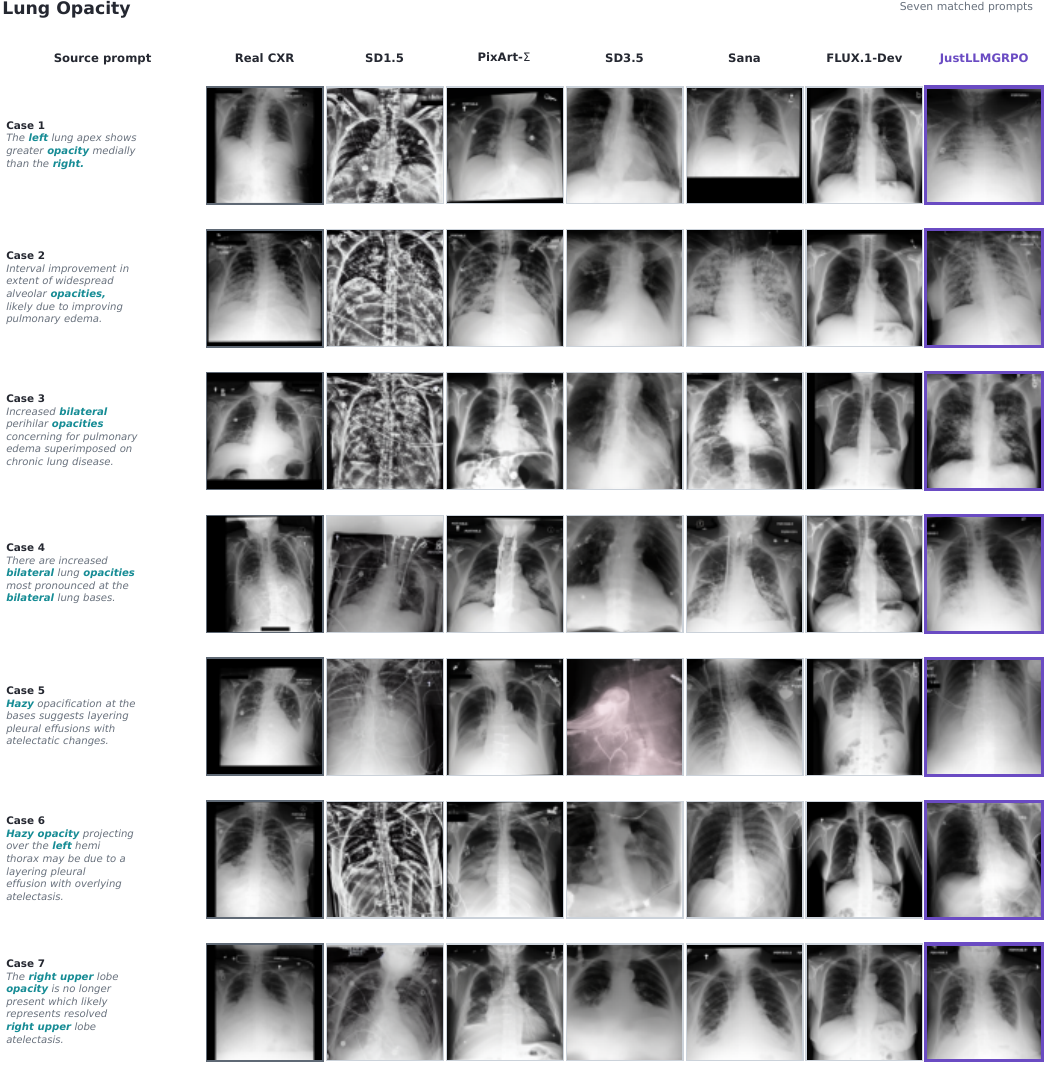}{Lung Opacity}{supp-lung-opacity}
\qualpage{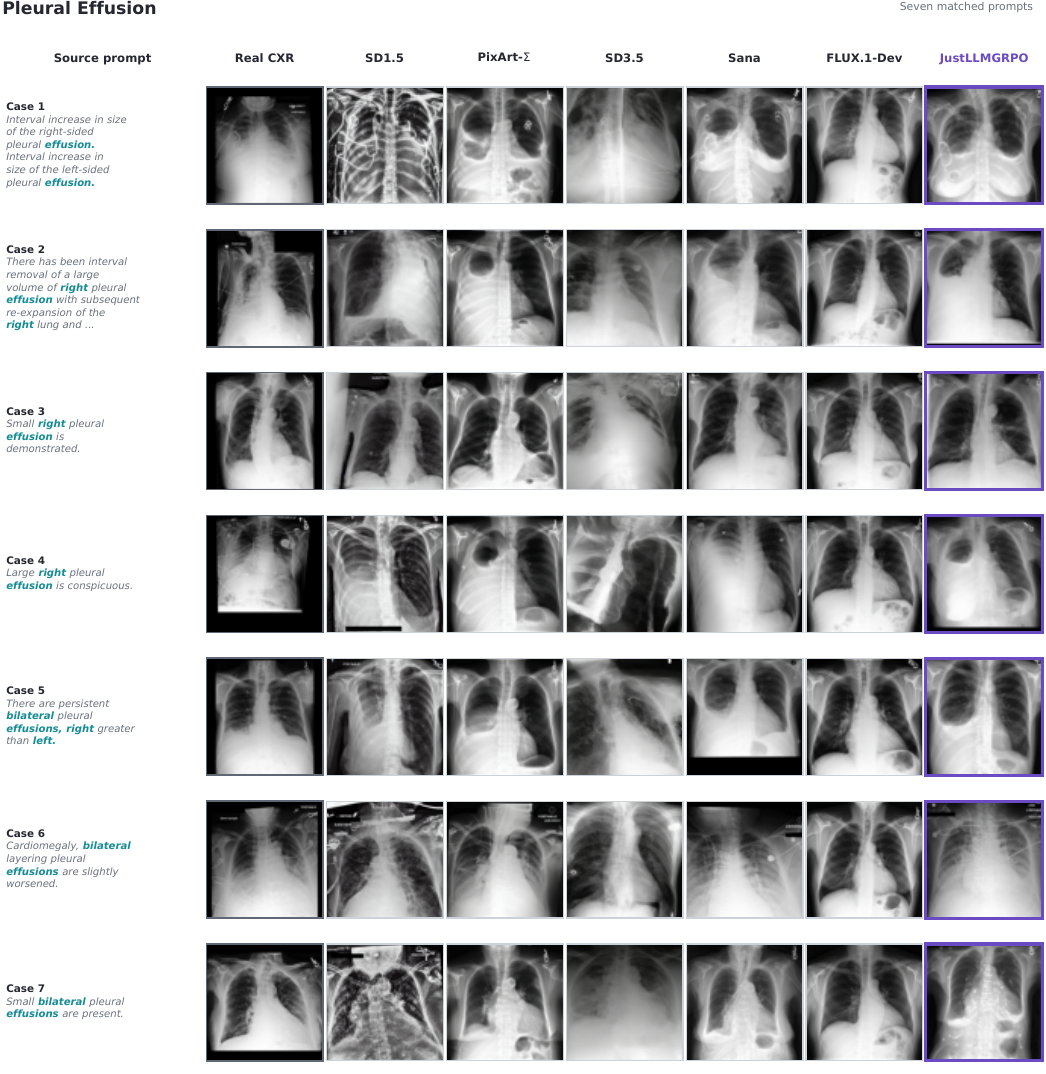}{Pleural Effusion}{supp-pleural-effusion}
\qualpage{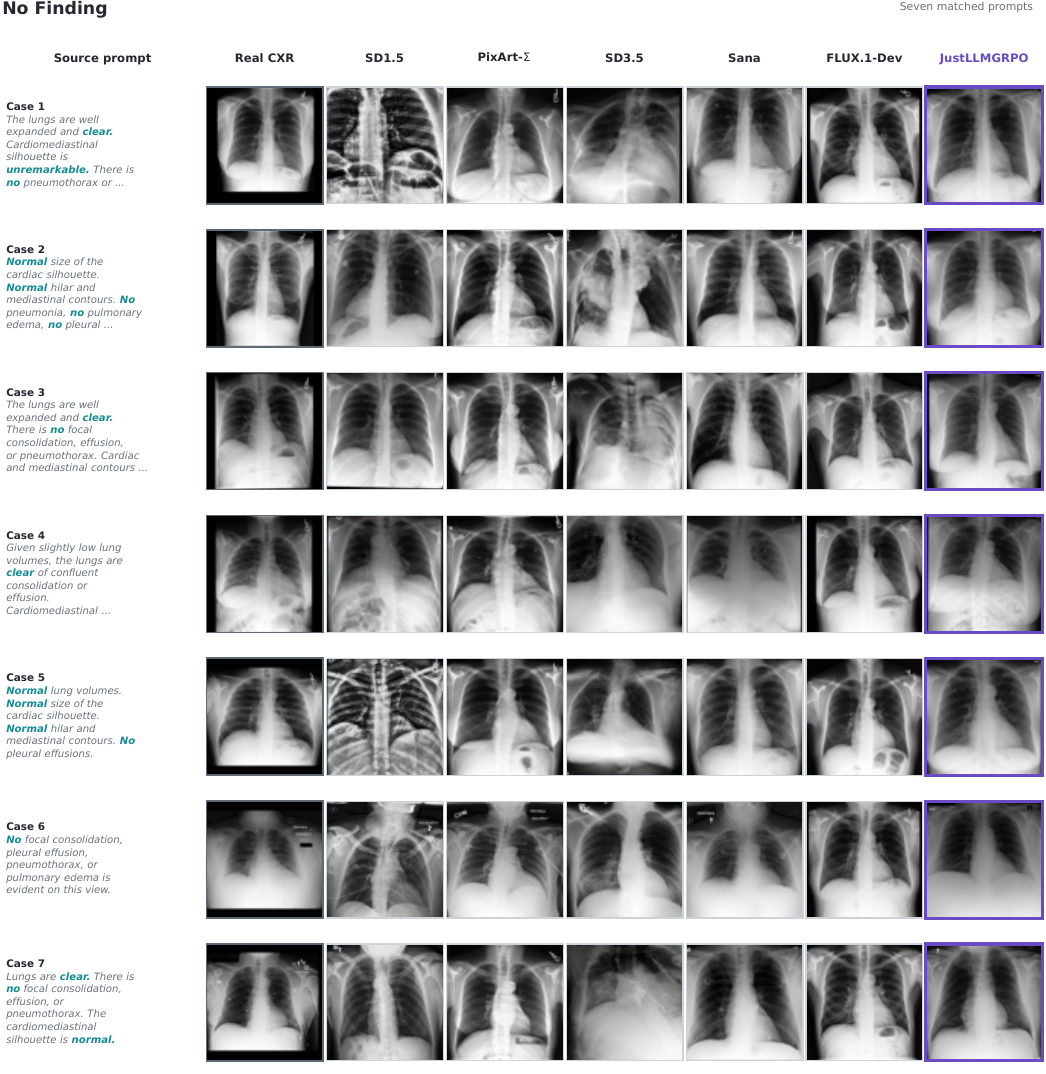}{No Finding}{supp-no-finding}
\qualpage{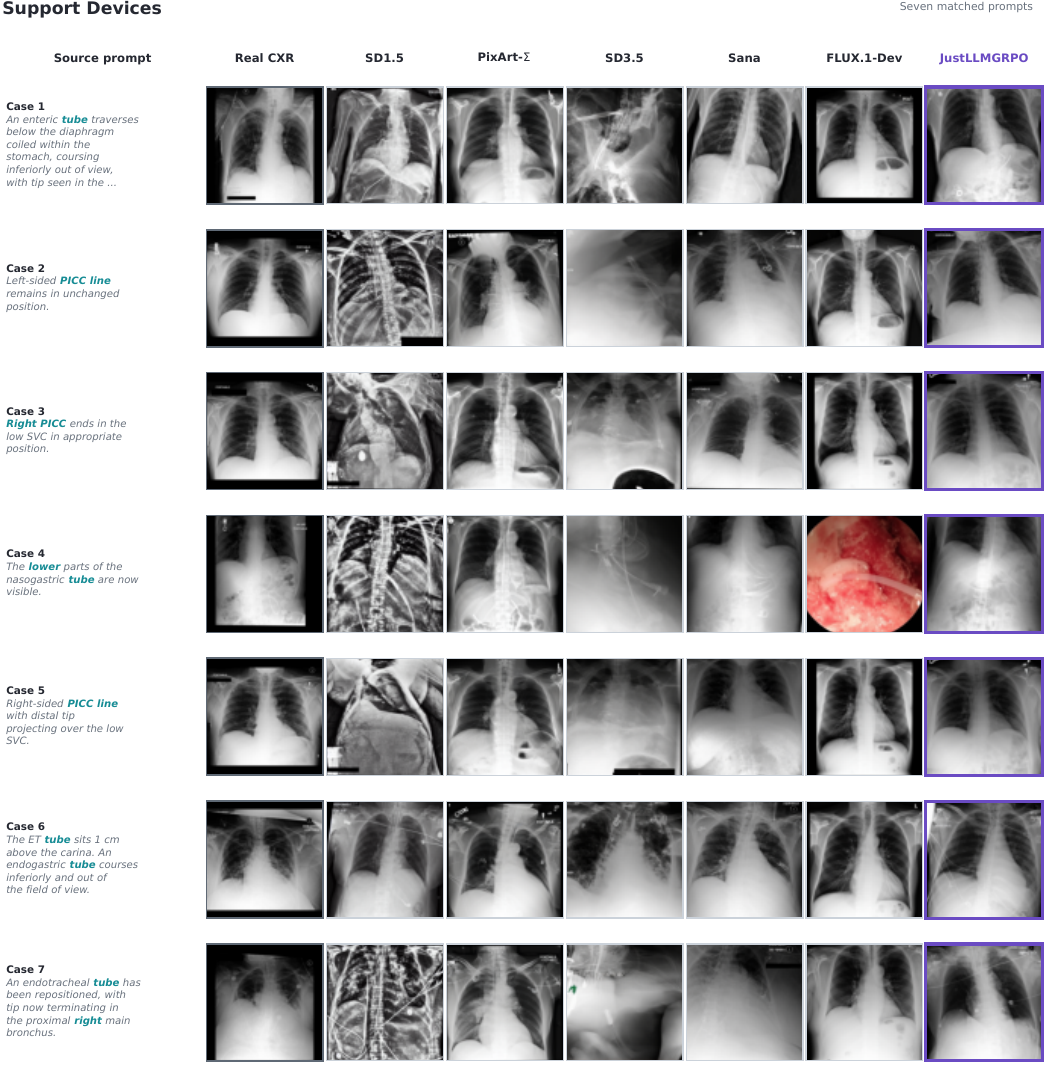}{Support Devices}{supp-support-devices}

\end{document}